\documentclass{article}

\usepackage[preprint]{neurips_2019}

\usepackage[utf8]{inputenc} 
\usepackage[T1]{fontenc}    
\usepackage{hyperref}       
\usepackage{url}            
\usepackage{booktabs}       
\usepackage{amsfonts}       
\usepackage{nicefrac}       
\usepackage{microtype}      
\usepackage{cite}           
\usepackage{graphicx}       
\usepackage{scalerel}       
\usepackage{xcolor}         
\usepackage{wrapfig}

\title{Adversarial Learning of Deepfakes in Accounting}

\author{%
  Marco Schreyer\\
  Institute of Computer Science\\
  University of St. Gallen\\
  \texttt{marco.schreyer@unisg.ch} \\
  \And
  Timur Sattarov \\
  Department of Statistics \\
  Deutsche Bundesbank \\
  \texttt{timur.sattarov@bundesbank.de} \\
  \AND
  Bernd Reimer \\
  Forensic Services \\
  PricewaterhouseCoopers GmbH \\
  \texttt{bernd.reimer@pwc.com} \\
  \And
  Damian Borth \\
  Institute of Computer Science \\
  University of St. Gallen \\
  \texttt{damian.borth@unisg.ch} \\
}

\begin{document}

\maketitle

\begin{abstract}
Nowadays, organizations collect vast quantities of accounting relevant transactions, referred to as 'journal entries', in 'Enterprise Resource Planning' (ERP) systems. The aggregation of those entries ultimately defines an organization's financial statement. To detect potential misstatements and fraud, international audit standards demand auditors to directly assess journal entries using 'Computer Assisted Audit Techniques' (CAATs). At the same time, discoveries in deep learning research revealed that machine learning models are vulnerable to 'adversarial attacks'. It also became evident that such attack techniques can be misused to generate 'deepfakes' designed to directly attack the perception of humans by creating convincingly altered media content. The research of such developments and their potential impact on the finance and accounting domain is still in its early stage. We believe that it is of vital relevance to investigate how such techniques could be maliciously misused in this sphere. In this work, we show an adversarial attack against CAATs using deep neural networks. We first introduce a real-world 'thread model' designed to camouflage accounting anomalies such as fraudulent journal entries. Second, we show that adversarial autoencoder neural networks are capable of learning a human interpretable model of journal entries that disentangles the entries latent generative factors. Finally, we demonstrate how such a model can be maliciously misused by a perpetrator to generate robust 'adversarial' journal entries that mislead CAATs.
\end{abstract}

\section{Introduction}
\label{sec:introduction}

Over the years, machine learning techniques, and in particular deep neural networks \cite{lecun2015}, created advances across a diverse range of application domains such as image classification \cite{krizhevsky2012}, speech recognition \cite{saon2015}, language translation \cite{sutskever2014} and game-play \cite{silver2017}. Due to their broad application, these developments also raised awareness of the unsafe aspects of machine learning, which become increasingly important. Intriguing discoveries in deep learning research revealed that a variety of machine learning models, even simple regression models, are vulnerable and exhibit 'intrinsic blind spots'. In computer vision, Szegedy et al. \cite{szegedy2013} and Goodfellow et al. \cite{goodfellow2014a} were among the first who demonstrated that small perturbations added to an image, resulted in misclassifications by machine learning models. Such perturbations, referred to as 'adversarial examples', pose a threat to a variety of real-world applications e.g. autonomous driving \cite{evtimov2017}, speech recognition \cite{alzantot2018}, text generation \cite{chen2018} or reinforcement learning \cite{huang2017}. 'Adversarial attacks' are deliberately designed to exploit such vulnerabilities and cause a machine learning model to make a mistake. Recently, another deep learning based attack type, referred to as 'deepfakes', gained considerable attention \cite{Mack2017}. Deepfakes denote convincingly manipulated media content, e.g., by altering its audio and video content. Most of the alterations make a person appear to say or do something that the person never said or did\footnote{BuzzFeedVideo, 'You Won’t Believe What Obama Says In This Video!', Published 04/17/2018, \newline https://www.youtube.com/watch?v=cQ54GDm1eL0.}.
Instead of attacking a machine learning model, deepfakes are designed to directly target a human viewers visual and acoustic perception. In the past, the creation of such altered media used to be reserved for a small group of highly trained professionals. With the advent of deep adversarial learning, it became broadly accessible within reach of almost any individual with a computer \cite{guera2018}. The early detection of such deepfakes is of high relevance in the context of societal disinformation and are of serious concern in democratic discourses \cite{Sullivan2019}. The research of the potential impact of adversarial attacks and deepfakes in finance and accounting is still in an early stage. However, we believe that it is of vital relevance to understand how adversarial deep learning techniques could be maliciously misused in this sphere. This holds in particular for the creation of adversarial journal entries to cover-up fraudulent\footnote{In accordance with \cite{garner2014} the term fraud here refers to the use of one’s occupation for personal enrichment through the deliberate misuse or misapplication of the using organization’s resources or assets, e.g. the booking of fictitious sales, fraudulent invoices, wrong recording of expenses.} activities that might remain undetected by state-of-the-art 'Computer Assisted Audit Techniques' (CAATs). 

\begin{figure*}[t!]
    \begin{center}
        \includegraphics[width=13.5cm,trim={0cm 0cm 0.5cm 0cm}, clip]{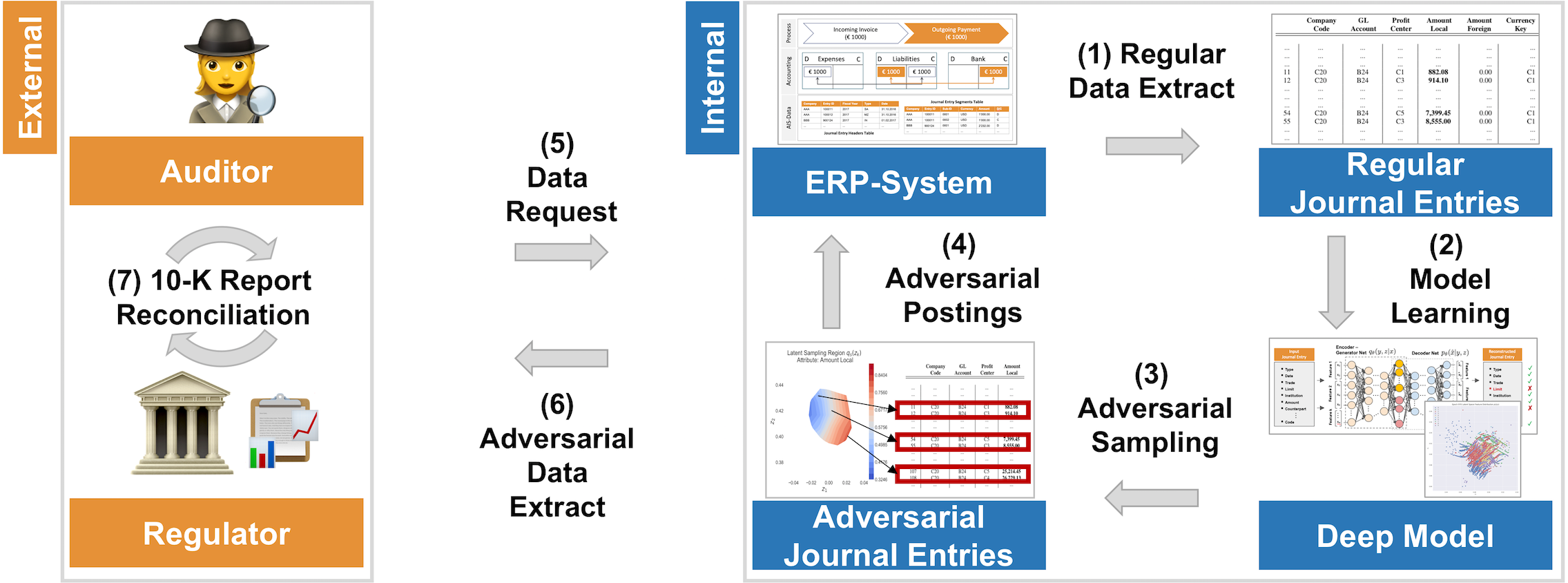}
    \end{center}
    \caption{Exemplary audit 'thread model' designed to camouflage accounting anomalies or fraudulent activities. Regular journal entries are replaced or enriched by the injection of 'fake' adversarial entries deliberately sampled from a deep generative model.}
    \label{fig:threadmodel}
\end{figure*}

Nowadays, organizations collect vast quantities of accounting relevant transactions referred to as 'journal entries' in 'Enterprise Resource Planning' (ERP) systems. The aggregation of those entries ultimately comprise an organization's financial statement. The trustworthiness of such financial statements plays a fundamental role in the economic decision making of investors and creditors \cite{IAS2007}. CAATs are typically designed to gather reasonable assurance that the financial statement of an organization is free from material misstatement ('true and fair presentation') and traces of fraud \cite{AICPA2002}, \cite{IFAC2009}. To detect potential fraudulent activities, international audit standards require the direct assessment of a financial statements underlying accounting records referred to as \textit{journal entries} \cite{CAQ2018}. The toolbox of applied CAATs, ranges from risk-oriented database queries \cite{coderre2009a}, \cite{coderre2009b} referred to as 'red-flag' tests (e.g. queries for postings outside the usual office hours, multiple vendor bank account changes, backdated expense account adjustments), over statistical and data mining based approaches \cite{bolton2002}, \cite{phua2010} (e.g. Benford’s Law \cite{Benford1938} or financial ratio analysis \cite{spathis2002}) to more recent deep learning inspired techniques \cite{schreyer2017} (e.g. autoencoder neural networks). However, even after uncovering a novel fraudulent 'modus operandi' using state-of-the-art CAATs, fraudsters continuously adapt until these techniques result in the generation of false negatives again. 

In the context of the rapid development of deep adversarial learning, this observation raises the following questions: Are financial audits vulnerable to adversarial attacks? And if so, are state-of-the-art CAATs able to detect such attacks? In this paper, we present a deep learning-based adversarial attack against CAATs. We regard this work to be an initial step towards the investigation of such future challenges in financial audits. In summary, we present the following contributions: (i) we describe a real-world 'thread model' designed to camouflage accounting anomalies or fraudulent activities using adversarial entries; (ii) we show that deep neural networks can learn a model of journal entries that semantically disentangles the entries latent generative factors; and, (iii) we demonstrate how such a learned model can be maliciously misused by a potential perpetrator to create 'adversarial' journal entries that are deliberately designed to misguide auditors CAATs during a financial audit.

\section{Related work}
\label{sec:related_work}

Due to its threatening nature, adversarial attacks triggered a sizable body of research by practitioners and academia \cite{akhtar2018}, \cite{chakraborty2018}. Several references propose detection and defense mechanisms to improve the robustness of machine learning models against such attacks \cite{madry2017}. In the realm of this work, we focus our literature survey on (i) adversarial attacks under real-world conditions, and (ii) deepfakes designed to attack a humans perception.

\textbf{Real World 'Adversarial-Attack' Techniques}: Kurakin et al. in \cite{kurakin2016} showed the robustness of adversarial attacks in a real-world setup. They printed adversarial images and took snapshots of the images using a cell-phone camera. It was shown that a significant fraction of the photos was miss-classified by trained models when perceived through the camera. Etimov et al. in \cite{evtimov2017} built on these attacks and studied distinct types of road-sign perturbations. They demonstrated that physical perturbations, e.g., in the form of stickers added to road-signs, resulted in failures of the road-sign recognition systems. Athalye et al. in \cite{athalye2017} introduced a method for construction 3D objects able to fool neural networks across a wide variety of angles and viewpoints. Papernot et al. in \cite{papernot2017} showed the first real-world black-box attacks against a set of remotely hosted neural network classifiers. Using 'gradient-masking' and attack calibration, the attacks resulted in a high fraction of miss-classifications by the targeted networks. Similarly, Liu et al. in \cite{liu2016} showed that an ensemble of targeted black-box attacks against an image classification web-service provider resulted in high rates of misclassifications. In \cite{grosse2016}, Grosse et al. successfully attacked neural networks used as malware classifiers with adversarial perturbed android applications. Melis et al. in \cite{melis2017} demonstrated the vulnerability of robots to adversarial manipulations of input images. Xu et al. in \cite{xu2017} showed that commonly used compositional and non-compositional 'Visual Question and Answer' architectures are vulnerable to adversarial attacks.

\textbf{Real World 'Deepfake' Techniques}: Korshunova et al. in \cite{korshunova2017} used neural style transfer \cite{gatys2016} to conduct face-swapping of celebrities. Their model was trained on a dataset consisting over 200,000 celebrity face images and yield photorealistic results. Natsume et al. in \cite{natsume2018} proposed an approach to fuse random latent representations of the face and the hair regions of various headshot input images. The results trained and evaluated on the same celebrity face dataset showed a high synthesis quality of the swapped faces. Van den Oord et al. in \cite{van2016} proposed the 'WaveNet' architecture, an autoregressive generative model, designed to learn realistic voice characteristics of 109 different speakers. Kobayashi et al. \cite{kobayashi2017} conditioned such autoregressive models on acoustic features determined by a Gaussian-Mixture Models to convert the features of a source speaker into those of a random target speaker. Zhou et al. in \cite{zhou2018} successfully trained, a special RNN architecture named 'SampleRNNs', to conduct intra-gender and cross-gender voice conversions of four speakers and demonstrated their naturalness. Chan et al. in \cite{chan2018} showed the application of generative models in transferring the motions of a person captured in a source videos to a target person. The motion transfer was achieved by enhancing the 'Pix2PixHD' architecture \cite{wang2018} and filming the target person to impose posses of professional dancers. Kim et al. in \cite{kim2018} presented a methodology to transfer the full 3D head position, head rotation, face expression, eye gaze, and the eye blinking from a source actor to a portrait video of a target actor. More recently, Fried et al. in \cite{fried2019} showed a method to produce realistic videos in which the original dialogue of the speaker has been modified. 

To the best of our knowledge, this work presents the first analysis on the generation of adversarial journal entries to attack real-world financial statement audits using deep neural networks.

\section{Audit Threat Model}
\label{threat_model}

Most journal entries recorded within an organizations’ Enterprise Resource Planing (ERP) system relate to regular day-to-day business activities, e.g., the posting of invoices, recording of payments, or the booking of asset depreciation.

\subsection{Classification of Audit Attacks}
\label{threat_model:attacks}

To conduct fraud, perpetrators need to deviate from the "normal". Such a deviating behavior will be recorded by a minimal number of journal entries and their respective attribute values. We refer to journal entries exhibiting such deviating attribute values as \textit{accounting anomalies}. In this initial work, we introduce two attack classes designed to camouflage accounting anomalies, namely (a) their \textit{replacement} and (b) their \textit{augmentation}. In the anomaly replacement scenario, anomalies are removed and replaced by several adversarial journal entries. The adversarial journal entries intend to camouflage the removed anomalous entries. This attack class is deliberately designed to target rule-based CAATs, e.g., to cover up the circumvention of an invoice approval border in the 'procure to pay' process. In the anomaly augmentation scenario anomalies are supplemented by a set of additional adversarial journal entries. The supplementary entries aim to let the anomalous entries appear normal and not 'isolated' in the entire population of journal entries. This attack class is designed to target statistical CAATs, e.g., to cover up rarely used general ledger accounts, user accounts, or document types. Both scenarios can be denoted as 'targeted' \cite{liu2016} 'black-box' \cite{papernot2017} attacks, where the adversary doesn't have access to the auditors CAATs internals and fraudulent entries should be miss-classified as false negatives. 

\subsection{Audit Attack Strategy}
\label{threat_model:strategy}

To establish such attacks, we introduce a threat model depicted in Fig. \ref{fig:threadmodel}, in which a person or group of people, referred to as the \textit{perpetrator}, within an organization intends to camouflage fraudulent activities: Based on its regular business activities, an organization records regular journal entries in its ERP-system. To initiate the attack, a perpetrator will query the ERP system to extract the journal entries recorded by the system (1). Afterward, the extracted entries are misused to learn a deep disentangled model of their latent generative factors (2). Such a model could be learned from a subset or all extracted entries, e.g., the latent generative factors of all invoice postings of particular purchasing department and fiscal year (3). The model learned, is then exploited to sample adversarial ('fake') journal entries. The sampled adversarial entries are deliberately designed to cover-up fraudulent activities (a) already recorded or (b) to be recorded by the system. Furthermore, the sampling is conditioned on the disentangled latent generative factors to decrease the risk of being detected by the auditors' CAATS. The sampled adversarial entries are then posted in the productive system to replace or augment the regular journal entries (4). During a financial audit, auditors usually request a set or a subset of journal entries subject to the scope of the audit (5). Following, the organization queries its ERP-System and provides the requested data including the (i) fraudulent and (ii) adversarial entries to the auditor (6). We refer to such a data extract, including the adversarial journal entries as \textit{adversarial extract}. To ensure the completeness of the data received, auditors usually reconcile the obtained data extract with publicly available information (7). For example, by comparing the trial balances of the journal entries with the corresponding 10-K report published by a regulatory body such as the 'Security and Exchange Commission' (SEC). However, due to the compliance of the adversarial journal entries with the learned generative latent factors, the entries exhibit a high likelihood to remain undetected by the auditors CAATs. 

\section{Adversarial Accounting Model}
\label{model_learning}

\begin{figure}[t!]
    \begin{center}
        \includegraphics[width=10.75cm, trim={0cm 0cm 0.5cm 0cm}, clip]{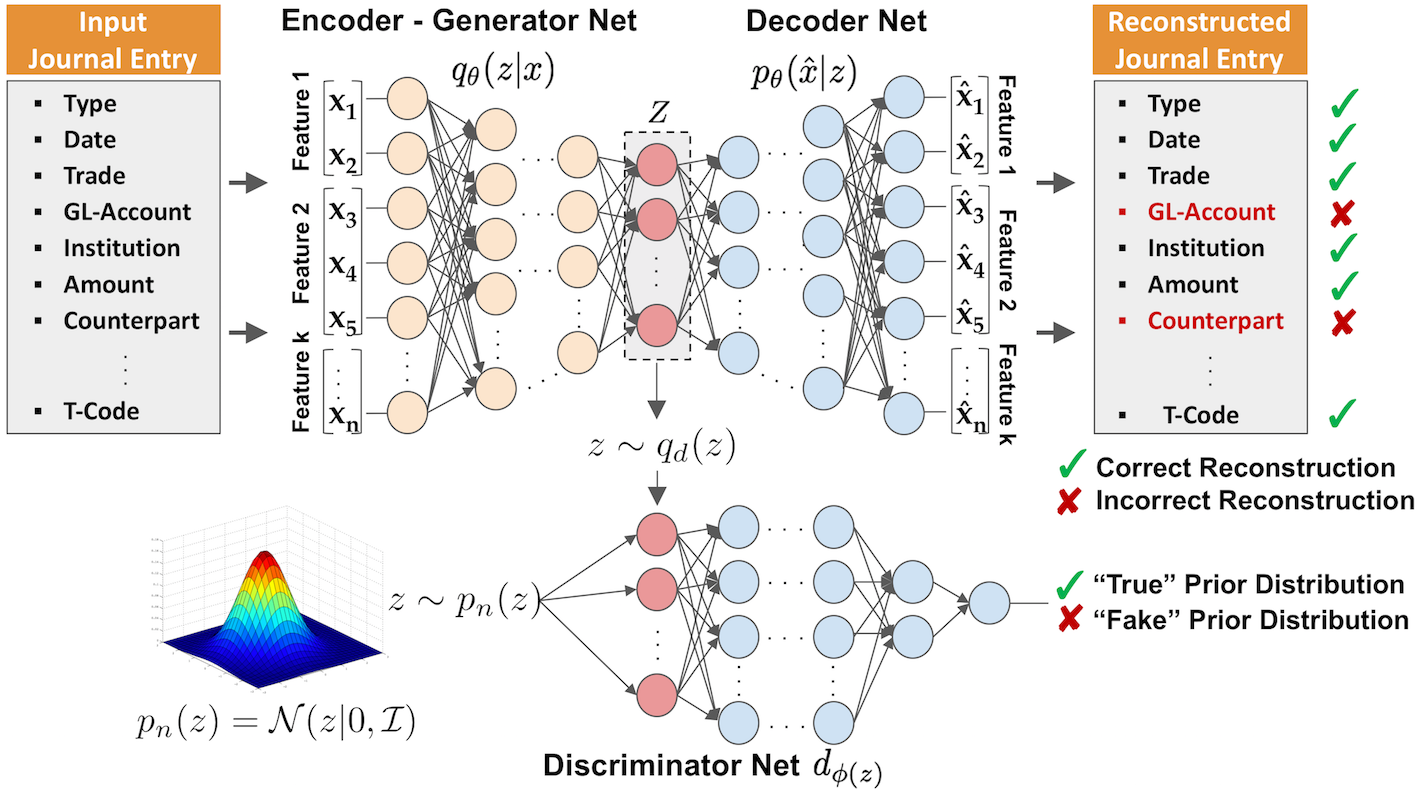}
    \end{center}
    \caption{The adversarial autoencoder architecture as introduced in \cite{makhzani2015}, applied to learn a disentangled and  human interpretable representation of the journal entries generative latent factors.}
    \label{fig:architecture}
\end{figure}


Let formally $X$ be a set of $N$ journal entries $x^{1}, x^{2}, ..., x^{n}$ where each journal entry $x^{i}$ consists of $K$ attributes $x_{1}^{i}, x_{2}^{i}, ..., x_{j}^{i}, ..., x_{k}^{i}$. Thereby, $x_{j}^{i}$ denotes the $j^{th}$ attribute of the $i^{th}$ journal entry. The individual attributes $x_{j}$ describe the journal entries accounting specific details, e.g., the entries fiscal year, posting type, posting date, amount, general-ledger. We hypothesize that such entries $X$ are generated by unobservable ('latent') factors of variation in $Z$ that can be recovered by an unsupervised deep learning algorithm \cite{bengio2013}. Similarly to \cite{bouchacourt2018} we also assume that such factors constitute a latent hierarchy. In this initial work, we distinguish between (a) \textit{high-order} and (b) \textit{low-order} latent factors of variation. High-order latent factors originate from the entries general semantics such as the underlying business process or organizational function, e.g., the variation of journal entry attributes when comparing vendor invoice postings to depreciation postings or customer payment postings. Low-order latent factors originate from the entries business transaction nuances or characteristics, e.g., the variations of journal entry attributes when comparing vendor invoice postings in terms of their posting time, the number of invoice line items, or the invoice amount. Directly calculating the posterior distribution $q_{d}(z)$ over the latent factors in $Z$ is often intractable \cite{kingma2013}. We aim to learn a model $q_{\theta}(z|x)$ of a given set of journal entries that approximates the intractable posterior. To achieve this, we use a deep neural network architecture referred to as Adversarial Autoencoder (AAE) introduced by Makhzani et al. in \cite{makhzani2015}. We impose a deliberately chosen prior $p_{n}(z)$ when training the AAE to obtain a human interpretable posterior distribution that disentangles the entries latent generative factors. Once such a model is learned, it provides a human interpretable 'blueprint' of how to control the generation of robust adversarial journal entries.

\subsection{Adversarial Autoencoder Networks}

The AAE architecture, as shown in Fig. \ref{fig:architecture}, extends the concept of Autoencoder Neural Networks (AENs) \cite{hinton2006} by imposing an arbitrary prior distribution $p_{n}(z)$ over the AENs latent factors in $Z$ using a Generative Adversarial Network (GAN) training setup \cite{goodfellow2014b}. The AAE training is conducted jointly in two phases, (a) a reconstruction phase and (b) an adversarial regularization phase executed on each mini-batch. In the reconstruction phase, the AAE's encoder network $q_{\theta}(z|x)$ is trained to approximate the intractable posterior of the entries latent generative factors. Given an arbitrary distribution of journal entries $p_{d}(x)$ the to be learned posterior $q_{d}(z)$ is formally defined as follows: 

\begin{equation}
    q_{d}(z) = \int_{x} q_{\theta}(z|x)p_{d}(x)\textit{d}x
\end{equation}

The AAE'S decoder network $p_{\theta}(\hat{x}|z)$ is trained to reconstruct each entry $\hat{x}^{i}$ based on its learned latent representations $z^{i} \in Z$ as faithfully as possible by minimizing the entry's reconstruction error. In the regularization phase, an adversarial training setup is applied, were the encoder $q_{\theta}(z|x)$ of the AAE functions as generator network. Thereby a discriminator network $d_{\phi}(z)$ is attached on top of the learned latent code vector $Z$. Similarly to GANs, the discriminator network of the AAE is trained to distinguish samples of an arbitrary imposed prior distribution $p_{n}(z)$ in $Z$ from the learned aggregated posterior distribution $q_{d}(z)$. In parallel, the encoder network is trained to learn a posterior distribution $p_{n}(z) \approx q_{d}(z)$ that fools the discriminator network into thinking that samples drawn from the posterior $q_{d}(z)$ originate from the imposed prior $p_{n}(z)$.

\subsection{Datasets and Data Preparation}

In general, SAP ERP systems record journal entries and their corresponding attributes predominantly in two database tables: (a) the table 'Accounting Document Headers' (technically: 'BKPF') contains the meta-information of a journal entry, such as document id, type, date, time, or currency, while (b) the table  'Accounting Document Segments' (technically: 'BSEG') contains the entry details, such as posting key, general ledger account, debit-credit information, or posting amount. In this work, we extract a subset of the most discriminative journal entry attributes of the BKPF and BSEG table. In our experiments, we use two datasets of journal entries: a real-world and a synthetic dataset referred to as \textit{Data-A} and \textit{Data-B} in the following. Data-A is an extract of an SAP ERP instance and contains a total of $307,457$ journal entry line items consisting of six categorical and two continuous attributes, denoted as $x_{cat}$ and $x_{con}$ respectively. The dataset encompasses the entire population of journal entries of a single fiscal year\footnote{In compliance with strict data privacy regulations, all journal entry attributes of Data-A have been anonymized using an irreversible one-way hash function during the data extraction process.}. Data-B is an excerpt of the synthetic dataset\footnote{The original dataset is publicly available via the Kaggle predictive modeling and analytics competitions platform and can be obtained using the following link: https://www.kaggle.com/ntnu-testimon/paysim1.} presented in \cite{paysim} and contains a total of $533,009$ journal entry line items.

\subsection{Adversarial Autoencoder Training}

Our architectural setup, shown in Fig. \ref{fig:architecture}, follows the AAE architecture described in \cite{makhzani2015}. In the training's reconstruction phase, we use a combined loss $\mathcal{L}^{\scaleto{RE}{3pt}}_{\theta}$ as proposed in \cite{schreyer2019}, when optimizing the parameters $\theta$ of the encoder $q_{\theta}(z|x)$ and decoder $p_{\theta}(\hat{x}|z)$, formally defined by:

\begin{equation}
    \mathcal{L}^{\scaleto{RE}{3pt}}_{\theta} = - \gamma \hspace{1mm} \frac{1}{N} \sum^{N-1}_{i=0} [x^{i}_{cat} \log(\hat{x}^{i}_{cat})] + (1 - \gamma) \hspace{1mm} \frac{1}{N} \sum^{N-1}_{i=0}[(x^{i}_{con}-\hat{x}^{i}_{con})^{2}]
\end{equation}

where $x^{i}=0,...,N-1 \sim q_{d}(x)$ and $N$ denotes the size of the training batch. The parameter $\gamma$ balances both, (a) the cross-entropy loss obtained for each entry's categorical attributes $\hat{x}_{cat}$ and (b) the mean-squared-error loss obtained for its continuous attributes $\hat{x}_{con}$. In the regularization phase, we optimize the adversarial loss $\mathcal{L}^{\scaleto{DI}{3pt}}_{\theta, \phi}$ as proposed in \cite{goodfellow2014b}, when optimizing the parameters $\theta$ of the generator $q_{\theta}(z|x)$ and the parameters $\phi$ of the discriminator $d_{\phi}(z)$, formally defined by:

\begin{equation}
    \mathcal{L}^{\scaleto{DI}{3pt}}_{\theta, \phi} = - \frac{1}{N} \sum^{N-1}_{i=0}[\log d_{\phi}(z^{i})] - \frac{1}{N} \sum^{2N}_{j=N}[\log(1-d_{\phi}(q_{\theta}(x^{i}))]
\end{equation}

where $z^{i}=0,...,N-1 \sim p_{n}(z)$ and $x^{i}=N,...,2N \sim p_{d}(x)$ and $N$ denotes the size of the training batch. We train the three networks that constitute the AAE in parallel with mini-batch wise stochastic gradient descent for max. of $10,000$ training epochs and apply early stopping once $\mathcal{L}_{\theta}^{\scaleto{RE}{3pt}}$ converges (architectural details and training parameters are outlined in the appendixes). To disentangle the journal entries generative latent factors, we sample from a prior distribution $p_{n}(z)$ that constitutes an equidistant grid of $\tau$ isotropic Gaussians denoted by $\mathcal{N}(\mu,\mathcal{I})$, where $\mu \in \mathbb{R}^{2}$. Figure \ref{fig:disentanglement} (left) shows an exemplary prior consisting of a 2D-grid of $\tau=25$ equidistant Gaussians.

\subsection{Latent Factor Disentanglement}

\begin{figure}[t!]
    \begin{center}
        \includegraphics[width=0.8\textwidth]{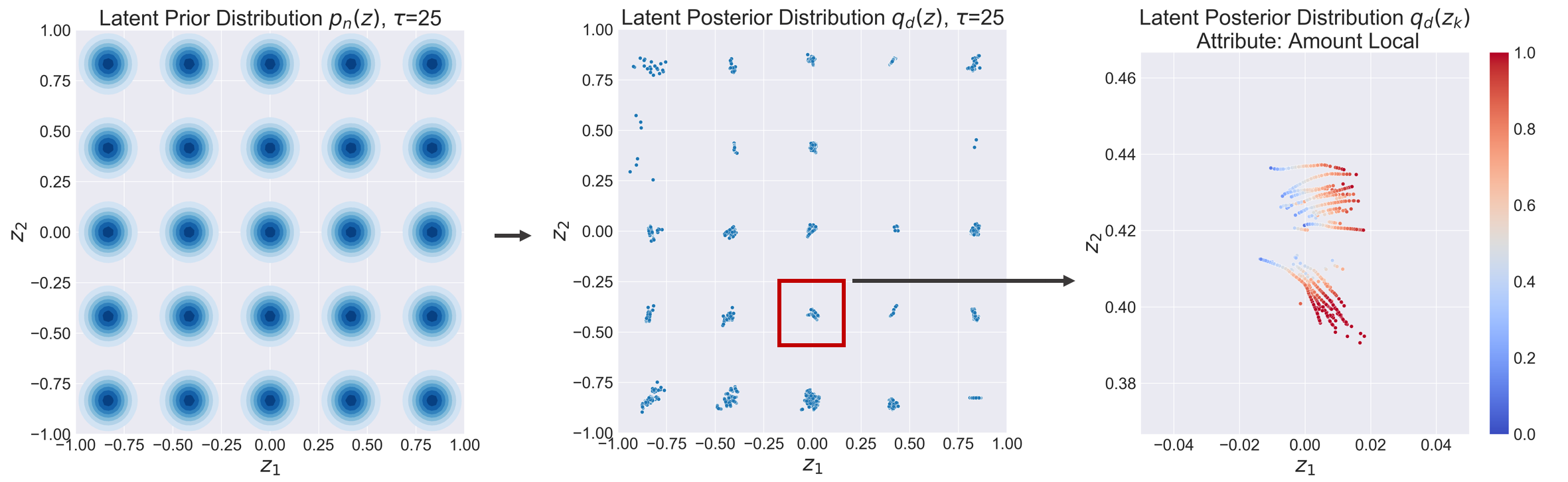}
    \end{center}
    \caption{Exemplary prior distribution $p_{n}(x)$ consisting of a. 2D-grid of $\tau=25$ equidistant Gaussians (left). Learned latent posterior distribution $q_{d}(z)$ that disentangles the journal entries high-order factors of variation into the distinct Gaussians (middle). Learned low-order disentanglement of the journal entries $log$-normalized local posting amounts of a single Gaussian (right).}
    \label{fig:disentanglement}
\end{figure}

Upon successful training, we qualitatively inspect the learned disentanglement of the high- and low-order generative latent factors learned, as represented in $q_{d}(z)$. First, we examine the disentanglement of the high-order generative factors by investigating the distinct Gaussians of the posterior. Figure \ref{fig:disentanglement} (middle) shows an exemplary learned posterior probability distribution of Data-A and its partitioning into $\tau=25$ Gaussians. The investigation revealed that each Gaussian $z_{k}$ and its underlying journal entry representations $z^{i}$ correspond to a general accounting process, e.g., (i) automated payment postings, (ii) incoming vendor invoices, or (iii) material movements. Second, we examine the disentanglement of the low-order generative factors by investigating the distribution of representations $z^{i}$ that constitute each latent Gaussian. Figure \ref{fig:disentanglement} (right) shows the disentangled distribution of the entries local posting amounts represented by a single Gaussian of Data-A. The investigation revealed that the distribution of each Gaussian disentangles the nuances of the general accounting processes, e.g., (i) posted general-ledger accounts, (ii) posting amounts, or (iii) distinct vendors.

\section{Creation of Adversarial Accounting Records}
\label{sample_generation}

\begin{figure}[t!]
    \begin{center}
        \includegraphics[width=1.0\textwidth]{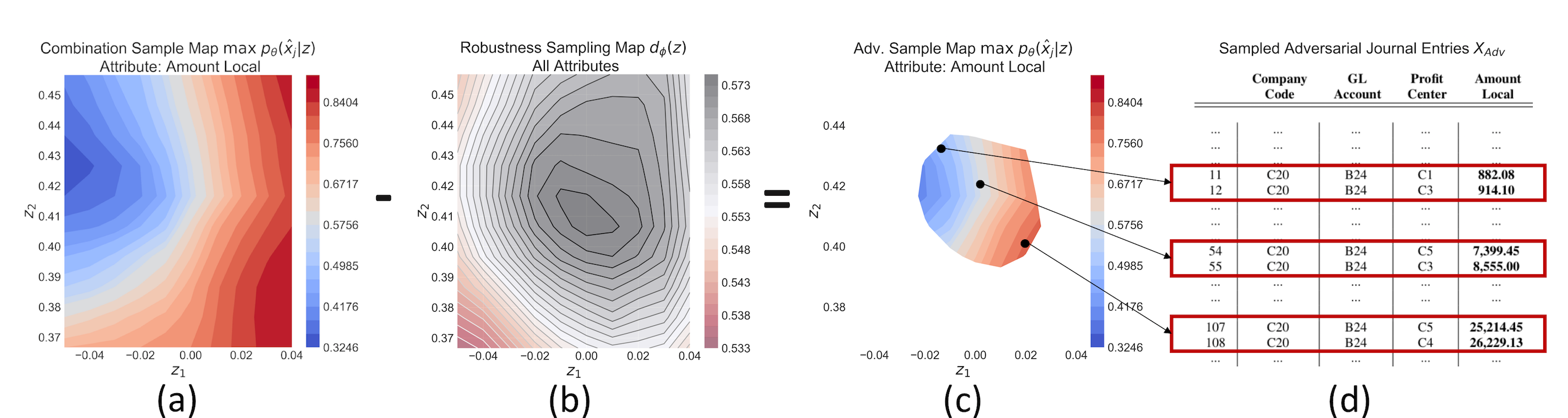}
    \end{center}
    \caption{Robust adversarial journal entry sampling: (a) the combination sampling map in $Z$ of the local posting amount attribute in Data-A, (b) the corresponding robustness sampling map, (c) the obtained adversarial sampling region $q_{s}(z_{k=14})$ combining (a) and (b) with $d_{\phi}(z) \geq 0.568$, and, (d) generated adversarial journal entries $X_{Adv}$ when sampling along the posting amount trajectory resulting in generated entries that exhibit an increased posting amount.}
    \label{fig:latentspace}
\end{figure}

To prepare for a potential sampling of adversarial journal entries, we conduct a detailed analysis of the models generative latent factors. Thereby, we sample equidistant in $Z$ with $z^{i} \in [-1,1]$ using an Euclidean distance of $\delta=10^{-4}$ between two neighboring samples $z^{i}$ and $z^{l}$ with $i \neq l$. For each sample we empirically investigate its attribute combination and attribute robustness:

\textbf{Attribute combination analysis:} First, we analyze from which region in $Z$ to sample to generate entries that exhibit a desired  attribute value combination characteristic, e.g., the chance of \textit{generating} a particular general-ledger, posting amount, and vendor combination. Given the AAE's decoder $p_{\theta}(\hat{x}|z)$, for each sample $z^{i} \in Z$ we derive its possibility of generating a particular attribute value $\hat{x}_{j}$. We are interested in likelihood trajectories where a traversal step $\delta$ in $Z$ results in a transition of one attribute value into another, formally denoted by $\max p_{\theta}(\hat{x}_{j}|z) \neq \max p_{\theta}(\hat{x}_{j}|z + \delta)$. Figure \ref{fig:latentspace} (a) shows the latent \textit{combination sample map} of the journal entries local posting amount attribute derived by the analysis of Data-A.

\textbf{Attribute robustness analysis:} Second, we analyze from which region in $Z$ to sample to generate entries that exhibit a desired attribute value combination robustness, e.g., the likelihood of \textit{observing} a specific general-ledger, posting amount, and vendor combination. Given the AAE's discriminator $d_{\phi}(z)$, for each sample $z^{i} \in Z$ we derive its likelihood of originating from the learned posterior. A high likelihood signifies the entry's compliance with the disentangled latent generative factors of the regular population of journal entries $X$. We are interested in likelihood trajectories where a traversal step $\delta$ in $Z$ results in a significant likelihood change $\rho$, formally denoted by $\|d_{\phi}(z + \delta)\| \geq d_{\phi}(z) + \rho$. Figure 4 (b) shows the latent \textit{robust sample map} derived by the analysis of Data-A.

Combining the both analysis results provides a 'blueprint' of controllable latent factors to generate robust adversarial journal entries. The objective of a perpetrators is thereby to sample from a latent space region $q_{s}(z_{k})$ to generates entries that exhibit a desired combination of fraudulent attribute values and a high attribute value robustness. Figure \ref{fig:latentspace} (c) illustrates such an exemplary adversarial sampling region $q_{s}(z_{k=14})$ obtained for the posting amount attribute in Data-A. Figure \ref{fig:latentspace} (d) depicts a set of generated adversarial entries when sampling along the posting amount trajectory in $q_{s}(z_{k=14})$. The generated entries comply with the regular entries characteristics and don't appear anomalous due to an unusual attribute value, e.g., a high posting amount, or an unusual attribute value combination, e.g., a posting amount and general-ledger that exhibit a low likelihood to co-occur in the set of regular journal entries. In the following, we describe two exemplary attack scenarios that illustrate the potential misuse of such a model and attack CAATs. Thereby, each example corresponds to one of the attack classes initially introduced:

An example of the \textit{anomaly replacement} attack scenario is illustrated in Tab. \ref{tab:replacement}. The to be replaced journal entry corresponds to an invoice posting exhibiting a total amount of \$47,632.45. The posting amount exceeds the organizations invoice approval border of \$25,0000 and may violate an internal control. To remain undetected by rule-based CAATs, five adversarial journal entries $X_{Adv}$ are sampled from the learned adversarial sampling region $q_{s}(z_{k=14})$ of Data-A (bottom). The sampled entries split the overall invoice amount into smaller amounts, each below the designated approval boarder. The sampling is achieved by traversing the latent factor trajectory of the amount attribute value while conditioning on the remaining attribute values of the to be replaced entry.

An example of the \textit{anomaly augmentation} attack scenario is illustrated in Tab. \ref{tab:enrichment}. The to be augmented journal entry corresponds to a single payment posting booked to general ledger account 'B24' (top). To remain undetected by statistical CAATs, several adversarial journal entries $X_{Adv}$ are sampled from $q_{s}(z_{k=14})$ of Data-A (bottom). The sampled entries augment the single payment posting and obfuscate the isolated posting not to be detected as an anomaly in terms of seldom booked general ledger accounts. The sampling is achieved by conditioning on the general ledger attribute while traversing the remaining latent factor trajectories of the entry to be augmented.

\begin{table}[t!]
    \caption{Anomaly replacement: Journal entry that corresponds to an invoice posting exhibiting an amount of \$47,632.45 (top). The posted amount exceeds the organizations approval boarder of \$25,000. Adversarial journal entries, sampled from the learned model of Data-A (bottom). The sampled journal entries split the overall invoice amount (varied attribute values in bold) such that the likelihood of an alert triggered by a CAATs decreases.}
  \scriptsize
  \begin{center}
      \begin{tabular}{r c|c|c|c|c|c|r|r|c|}
        \multicolumn{1}{r}{\small Original }
        & \multicolumn{1}{c}{\textbf{ }}
        & \multicolumn{1}{c}{\textbf{ Company }}
        & \multicolumn{1}{c}{\textbf{ Posting }} 
        & \multicolumn{1}{c}{\textbf{ Account }}
        & \multicolumn{1}{c}{\textbf{ GL }}
        & \multicolumn{1}{c}{\textbf{ Profit }}
        & \multicolumn{1}{c}{\textbf{ Amount }}
        & \multicolumn{1}{c}{\textbf{ ... }}
        & \multicolumn{1}{c}{\textbf{ Currency }}\\
        \multicolumn{1}{r}{ \small entry: }
        & \multicolumn{1}{c}{\textbf{ }}
        & \multicolumn{1}{c}{\textbf{ Code }}
        & \multicolumn{1}{c}{\textbf{ Key }} 
        & \multicolumn{1}{c}{\textbf{ Key }}
        & \multicolumn{1}{c}{\textbf{ Account }}
        & \multicolumn{1}{c}{\textbf{ Center }}
        & \multicolumn{1}{c}{\textbf{ Local }}
        & \multicolumn{1}{c}{\textbf{ }}
        & \multicolumn{1}{c}{\textbf{ Key }}\\
        \cmidrule{2-10}\morecmidrules\cmidrule{2-10}
          & 1 & C20 & A1 & C1 & B1 & C20 & 47,632.45 & ... & C7 \\
        \multicolumn{1}{c}{}&\multicolumn{1}{c}{}&\multicolumn{1}{c}{}\\
       \end{tabular}
    \begin{tabular}{r c|c|c|c|c|c|r|r|c|}
        \multicolumn{1}{r}{\small Generated }
        & \multicolumn{1}{c}{\textbf{ }}
        & \multicolumn{1}{c}{\textbf{ Company }}
        & \multicolumn{1}{c}{\textbf{ Posting }} 
        & \multicolumn{1}{c}{\textbf{ Account }}
        & \multicolumn{1}{c}{\textbf{ GL }}
        & \multicolumn{1}{c}{\textbf{ Profit }}
        & \multicolumn{1}{c}{\textbf{ Amount }}
        & \multicolumn{1}{c}{\textbf{ ... }}
        & \multicolumn{1}{c}{\textbf{ Currency }}\\
        \multicolumn{1}{r}{\small entries: }
        & \multicolumn{1}{c}{\textbf{ }}
        & \multicolumn{1}{c}{\textbf{ Code }}
        & \multicolumn{1}{c}{\textbf{ Key }} 
        & \multicolumn{1}{c}{\textbf{ Key }}
        & \multicolumn{1}{c}{\textbf{ Account }}
        & \multicolumn{1}{c}{\textbf{ Center }}
        & \multicolumn{1}{c}{\textbf{ Local }}
        & \multicolumn{1}{c}{\textbf{ }}
        & \multicolumn{1}{c}{\textbf{ Key }}\\
        \cmidrule{2-10}\morecmidrules\cmidrule{2-10}
          & 1 & C20 & A1 & C1 & B1 & C20 & \textbf{2,381.62} & \textbf{...} & C7\\
          & 2 & C20 & A1 & C1 & B1 & C20 & \textbf{4,763.25} & \textbf{...} & C7\\
          & 3 & C20 & A1 & C1 & B1 & C20 & \textbf{11,908.11} & \textbf{...} & C7\\
          & 4 & C20 & A1 & C1 & B1 & C20 & \textbf{9,526.49} & \textbf{...} & C7\\
          & 5 & C20 & A1 & C1 & B1 & C20 & \textbf{19,052,98} & \textbf{...} & C7\\
    \end{tabular}
    \label{tab:replacement}
  \end{center}
\end{table}

\begin{table}[t!]
    \caption{Anomaly augmentation: Journal entry that corresponds to payment posting booked to seldom used general ledger account 'B24' (top). Adversarial journal entries, sampled from the learned model of Data-A (bottom). The sampled entries augment the single payment posting, and obfuscate the isolated posting (not varied attribute values in bold) such that the likelihood of an alert triggered by a CAATs decreases.}
  \scriptsize
  \begin{center}
      \begin{tabular}{r c|c|c|c|c|c|r|r|c|}
        \multicolumn{1}{r}{\small Original }
        & \multicolumn{1}{c}{\textbf{ }}
        & \multicolumn{1}{c}{\textbf{ Company }}
        & \multicolumn{1}{c}{\textbf{ Posting }} 
        & \multicolumn{1}{c}{\textbf{ Account }}
        & \multicolumn{1}{c}{\textbf{ GL }}
        & \multicolumn{1}{c}{\textbf{ Profit }}
        & \multicolumn{1}{c}{\textbf{ Amount }}
        & \multicolumn{1}{c}{\textbf{ ... }}
        & \multicolumn{1}{c}{\textbf{ Currency }}\\
        \multicolumn{1}{r}{ \small entry: }
        & \multicolumn{1}{c}{\textbf{ }}
        & \multicolumn{1}{c}{\textbf{ Code }}
        & \multicolumn{1}{c}{\textbf{ Key }} 
        & \multicolumn{1}{c}{\textbf{ Key }}
        & \multicolumn{1}{c}{\textbf{ Account }}
        & \multicolumn{1}{c}{\textbf{ Center }}
        & \multicolumn{1}{c}{\textbf{ Local }}
        & \multicolumn{1}{c}{\textbf{ }}
        & \multicolumn{1}{c}{\textbf{ Key }}\\
        \cmidrule{2-10}\morecmidrules\cmidrule{2-10}
          & 1 & C20 & A2 & C2 & B24 & C1 & 8,920.00 & ... & C1\\
        \multicolumn{1}{c}{}&\multicolumn{1}{c}{}&\multicolumn{1}{c}{}\\
       \end{tabular}
    \begin{tabular}{r c|c|c|c|c|c|r|r|c|}
        \multicolumn{1}{r}{ \small Generated }
        & \multicolumn{1}{c}{\textbf{ }}
        & \multicolumn{1}{c}{\textbf{ Company }}
        & \multicolumn{1}{c}{\textbf{ Posting }} 
        & \multicolumn{1}{c}{\textbf{ Account }}
        & \multicolumn{1}{c}{\textbf{ GL }}
        & \multicolumn{1}{c}{\textbf{ Profit }}
        & \multicolumn{1}{c}{\textbf{ Amount }}
        & \multicolumn{1}{c}{\textbf{ ... }}
        & \multicolumn{1}{c}{\textbf{ Currency }}\\
        \multicolumn{1}{r}{ \small entries: }
        & \multicolumn{1}{c}{\textbf{ }}
        & \multicolumn{1}{c}{\textbf{ Code }}
        & \multicolumn{1}{c}{\textbf{ Key }} 
        & \multicolumn{1}{c}{\textbf{ Key }}
        & \multicolumn{1}{c}{\textbf{ Account }}
        & \multicolumn{1}{c}{\textbf{ Center }}
        & \multicolumn{1}{c}{\textbf{ Local }}
        & \multicolumn{1}{c}{\textbf{ }}
        & \multicolumn{1}{c}{\textbf{ Key }}\\
        \cmidrule{2-10}\morecmidrules\cmidrule{2-10}
          & 1 & C20 & A2 & C2 & \textbf{B24} & C1 & 8,082.08 & ... & C1\\
          & 2 & C20 & A2 & C2 & \textbf{B24} & C3 & 9,132.10 & ... & C1\\
          & ... & ... & ... & ... & ... & ... & ... & ... & ...\\
          & 14 & C20 & A2 & C2 & \textbf{B24} & C5 & 7,399.45 & ... & C1\\
          & 15 & C20 & A2 & C2 & \textbf{B24} & C3 & 8,555.00 & ... & C1\\
          & ... & ... & ... & ... & ... & ... & ... & ... & ...\\
    \end{tabular}
    \label{tab:enrichment}
  \end{center}
\end{table}

\section{Conclusion}

In this work, we conducted a first analysis of the potential impact of 'deepfake' accounting records in the context of financial audits. We introduced a thread model to camouflage fraudulent activities in real-world journal entry data recorded in ERP systems. We demonstrated that deep neural networks can be trained in an adversarial setup to disentangle the entries underlying latent generative factors. We also provided initial evidence that such a model of disentangled latent generative factors can be maliciously misused by a potential perpetrator to attack CAATs regularly applied in financial audits. 

\section*{Acknowledgements}

We thank the members of the statistics department at Deutsche Bundesbank for their valuable review and remarks. Opinions expressed in this work are solely those of the authors, and do not necessarily reflect the view of the Deutsche Bundesbank nor PricewaterhouseCoopers (PwC) International Ltd. and its network firms.

\bibliographystyle{abbrv}
\bibliography{library}

\begin{thebibliography}{10}

\bibitem{AICPA2002}
AICPA.
\newblock {\em {AU Section 316 - Consideration of Fraud in a Financial
  Statement Audit}}.
\newblock American Institute of Certified Public Accountants (AICPA), 2002.

\bibitem{akhtar2018}
N.~Akhtar and A.~Mian.
\newblock Threat of adversarial attacks on deep learning in computer vision: A
  survey.
\newblock {\em IEEE Access}, 6:14410--14430, 2018.

\bibitem{alzantot2018}
M.~Alzantot, B.~Balaji, and M.~Srivastava.
\newblock Did you hear that? adversarial examples against automatic speech
  recognition.
\newblock {\em arXiv preprint arXiv:1801.00554}, 2018.

\bibitem{arjovsky2017}
M.~Arjovsky, S.~Chintala, and L.~Bottou.
\newblock Wasserstein gan.
\newblock {\em arXiv preprint arXiv:1701.07875}, 2017.

\bibitem{athalye2017}
A.~Athalye, L.~Engstrom, A.~Ilyas, and K.~Kwok.
\newblock Synthesizing robust adversarial examples.
\newblock {\em arXiv preprint arXiv:1707.07397}, 2017.

\bibitem{Benford1938}
F.~Benford.
\newblock {The Law of Anomalous Numbers}.
\newblock {\em Proceedings of the American Philosophical Society},
  78(4):551--572, 1938.

\bibitem{bengio2013}
Y.~Bengio, A.~Courville, and P.~Vincent.
\newblock Representation learning: A review and new perspectives.
\newblock {\em IEEE transactions on pattern analysis and machine intelligence},
  35(8):1798--1828, 2013.

\bibitem{bolton2002}
R.~J. Bolton and D.~J. Hand.
\newblock Statistical fraud detection: A review.
\newblock {\em Statistical science}, pages 235--249, 2002.

\bibitem{bouchacourt2018}
D.~Bouchacourt, R.~Tomioka, and S.~Nowozin.
\newblock Multi-level variational autoencoder: Learning disentangled
  representations from grouped observations.
\newblock In {\em Thirty-Second AAAI Conference on Artificial Intelligence},
  2018.

\bibitem{CAQ2018}
CAQ.
\newblock {\em {Practice Aid for Testing Journal Entries and Other Adjustments
  Pursuant to AU Section 316}}.
\newblock Center for Audit Quality, 2008.

\bibitem{chakraborty2018}
A.~Chakraborty, M.~Alam, V.~Dey, A.~Chattopadhyay, and D.~Mukhopadhyay.
\newblock Adversarial attacks and defences: A survey.
\newblock {\em arXiv preprint arXiv:1810.00069}, 2018.

\bibitem{chan2018}
C.~Chan, S.~Ginosar, T.~Zhou, and A.~A. Efros.
\newblock Everybody dance now.
\newblock {\em arXiv preprint arXiv:1808.07371}, 2018.

\bibitem{chen2018}
L.~Chen, S.~Dai, C.~Tao, H.~Zhang, Z.~Gan, D.~Shen, Y.~Zhang, G.~Wang,
  R.~Zhang, and L.~Carin.
\newblock Adversarial text generation via feature-mover's distance.
\newblock In {\em Advances in Neural Information Processing Systems}, pages
  4671--4682, 2018.

\bibitem{coderre2009b}
D.~Coderre.
\newblock {\em Computer-aided Fraud Prevention and Detection: A Step-by-step
  Guide}.
\newblock John Wiley \& Sons, 2009.

\bibitem{coderre2009a}
D.~Coderre.
\newblock {\em Fraud Analysis Techniques Using ACL}.
\newblock John Wiley \& Sons, 2009.

\bibitem{evtimov2017}
I.~Evtimov, K.~Eykholt, E.~Fernandes, T.~Kohno, B.~Li, A.~Prakash, A.~Rahmati,
  and D.~Song.
\newblock Robust physical-world attacks on deep learning models.
\newblock {\em arXiv preprint arXiv:1707.08945}, 1:1, 2017.

\bibitem{fried2019}
O.~Fried, A.~Tewari, M.~Zollh{\"o}fer, A.~Finkelstein, E.~Shechtman, D.~B.
  Goldman, K.~Genova, Z.~Jin, C.~Theobalt, and M.~Agrawala.
\newblock Text-based editing of talking-head video.
\newblock {\em arXiv preprint arXiv:1906.01524}, 2019.

\bibitem{garner2014}
B.~Garner.
\newblock {\em Black's Law Dictionary}.
\newblock Thomson Reuters, 2014.

\bibitem{gatys2016}
L.~A. Gatys, A.~S. Ecker, and M.~Bethge.
\newblock Image style transfer using convolutional neural networks.
\newblock In {\em Proceedings of the IEEE conference on computer vision and
  pattern recognition}, pages 2414--2423, 2016.

\bibitem{glorot2010}
X.~Glorot and Y.~Bengio.
\newblock Understanding the difficulty of training deep feedforward neural
  networks.
\newblock In {\em Proceedings of the thirteenth international conference on
  artificial intelligence and statistics}, pages 249--256, 2010.

\bibitem{goodfellow2014b}
I.~Goodfellow, J.~Pouget-Abadie, M.~Mirza, B.~Xu, D.~Warde-Farley, S.~Ozair,
  A.~Courville, and Y.~Bengio.
\newblock Generative adversarial nets.
\newblock In {\em Advances in neural information processing systems}, pages
  2672--2680, 2014.

\bibitem{goodfellow2014a}
I.~J. Goodfellow, J.~Shlens, and C.~Szegedy.
\newblock Explaining and harnessing adversarial examples.
\newblock {\em arXiv preprint arXiv:1412.6572}, 2014.

\bibitem{grosse2016}
K.~Grosse, N.~Papernot, P.~Manoharan, M.~Backes, and P.~McDaniel.
\newblock Adversarial perturbations against deep neural networks for malware
  classification.
\newblock {\em arXiv preprint arXiv:1606.04435}, 2016.

\bibitem{guera2018}
D.~G{\"u}era and E.~J. Delp.
\newblock Deepfake video detection using recurrent neural networks.
\newblock In {\em 2018 15th IEEE International Conference on Advanced Video and
  Signal Based Surveillance (AVSS)}, pages 1--6. IEEE, 2018.

\bibitem{hinton2006}
G.~E. Hinton and R.~R. Salakhutdinov.
\newblock Reducing the dimensionality of data with neural networks.
\newblock {\em science}, 313(5786):504--507, 2006.

\bibitem{huang2017}
S.~Huang, N.~Papernot, I.~Goodfellow, Y.~Duan, and P.~Abbeel.
\newblock Adversarial attacks on neural network policies.
\newblock {\em arXiv preprint arXiv:1702.02284}, 2017.

\bibitem{IAS2007}
IFAC.
\newblock {\em {International Accounting Standard (ISA) 1 - Presentation of
  Financial Statements}}.
\newblock International Federation of Accountants (IFAC), 2007.

\bibitem{IFAC2009}
IFAC.
\newblock {\em {International Standards on Auditing 240, The Auditor's
  Responsibilities Relating to Fraud in an Audit of Financial Statements}}.
\newblock International Federation of Accountants (IFAC), 2009.

\bibitem{kim2018}
H.~Kim, P.~Carrido, A.~Tewari, W.~Xu, J.~Thies, M.~Niessner, P.~P{\'e}rez,
  C.~Richardt, M.~Zollh{\"o}fer, and C.~Theobalt.
\newblock Deep video portraits.
\newblock {\em ACM Transactions on Graphics (TOG)}, 37(4):163, 2018.

\bibitem{kingma2014}
D.~P. Kingma and J.~Ba.
\newblock Adam: A method for stochastic optimization.
\newblock {\em arXiv preprint arXiv:1412.6980}, 2014.

\bibitem{kingma2013}
D.~P. Kingma and M.~Welling.
\newblock Auto-encoding variational bayes.
\newblock {\em arXiv preprint arXiv:1312.6114}, 2013.

\bibitem{kobayashi2017}
K.~Kobayashi, T.~Hayashi, A.~Tamamori, and T.~Toda.
\newblock Statistical voice conversion with wavenet-based waveform generation.
\newblock In {\em Interspeech}, pages 1138--1142, 2017.

\bibitem{korshunova2017}
I.~Korshunova, W.~Shi, J.~Dambre, and L.~Theis.
\newblock Fast face-swap using convolutional neural networks.
\newblock In {\em Proceedings of the IEEE International Conference on Computer
  Vision}, pages 3677--3685, 2017.

\bibitem{krizhevsky2012}
A.~Krizhevsky, I.~Sutskever, and G.~E. Hinton.
\newblock Imagenet classification with deep convolutional neural networks.
\newblock In {\em Advances in neural information processing systems}, pages
  1097--1105, 2012.

\bibitem{kurakin2016}
A.~Kurakin, I.~Goodfellow, and S.~Bengio.
\newblock Adversarial examples in the physical world.
\newblock {\em arXiv preprint arXiv:1607.02533}, 2016.

\bibitem{lecun2015}
Y.~LeCun, Y.~Bengio, and G.~Hinton.
\newblock Deep learning.
\newblock {\em Nature}, 521(7553):436, 2015.

\bibitem{liu2016}
Y.~Liu, X.~Chen, C.~Liu, and D.~Song.
\newblock Delving into transferable adversarial examples and black-box attacks.
\newblock {\em arXiv preprint arXiv:1611.02770}, 2016.

\bibitem{paysim}
E.~A. Lopez-Rojas, A.~Elmir, and S.~Axelsson.
\newblock Paysim: A financial mobile money simulator for fraud detection.
\newblock In {\em The 28th European Modeling and Simulation Symposium-EMSS,
  Larnaca, Cyprus}, 2016.

\bibitem{Mack2017}
D.~Mack.
\newblock This psa about fake news from barack obama is not what it appears.
\newblock {\em BuzzFeed.News}, 2016.

\bibitem{madry2017}
A.~Madry, A.~Makelov, L.~Schmidt, D.~Tsipras, and A.~Vladu.
\newblock Towards deep learning models resistant to adversarial attacks.
\newblock {\em arXiv preprint arXiv:1706.06083}, 2017.

\bibitem{makhzani2015}
A.~Makhzani, J.~Shlens, N.~Jaitly, I.~Goodfellow, and B.~Frey.
\newblock {Adversarial Autoencoders}.
\newblock {\em arXiv}, pages 1--10, 2015.

\bibitem{melis2017}
M.~Melis, A.~Demontis, B.~Biggio, G.~Brown, G.~Fumera, and F.~Roli.
\newblock Is deep learning safe for robot vision.
\newblock {\em Adversarial examples against the iCub humanoid. CoRR,
  abs/1708.06939}, 2017.

\bibitem{natsume2018}
R.~Natsume, T.~Yatagawa, and S.~Morishima.
\newblock Rsgan: face swapping and editing using face and hair representation
  in latent spaces.
\newblock {\em arXiv preprint arXiv:1804.03447}, 2018.

\bibitem{Sullivan2019}
D.~O'Sullivan.
\newblock When seeing is no longer believing.
\newblock {\em CNN Business}, 2017.

\bibitem{papernot2017}
N.~Papernot, P.~McDaniel, I.~Goodfellow, S.~Jha, Z.~B. Celik, and A.~Swami.
\newblock Practical black-box attacks against machine learning.
\newblock In {\em Proceedings of the 2017 ACM on Asia conference on computer
  and communications security}, pages 506--519. ACM, 2017.

\bibitem{paszke2017}
A.~Paszke, S.~Gross, S.~Chintala, G.~Chanan, E.~Yang, Z.~DeVito, Z.~Lin,
  A.~Desmaison, L.~Antiga, and A.~Lerer.
\newblock Automatic differentiation in pytorch.
\newblock {\em https://pytorch.org}, 2017.

\bibitem{phua2010}
C.~Phua, V.~Lee, K.~Smith, and R.~Gayler.
\newblock A comprehensive survey of data mining-based fraud detection research.
\newblock {\em arXiv preprint arXiv:1009.6119}, 2010.

\bibitem{saon2015}
G.~Saon, H.-K.~J. Kuo, S.~Rennie, and M.~Picheny.
\newblock The ibm 2015 english conversational telephone speech recognition
  system.
\newblock {\em arXiv preprint arXiv:1505.05899}, 2015.

\bibitem{schreyer2017}
M.~Schreyer, T.~Sattarov, D.~Borth, A.~Dengel, and B.~Reimer.
\newblock Detection of anomalies in large scale accounting data using deep
  autoencoder networks.
\newblock {\em arXiv preprint arXiv:1709.05254}, 2017.

\bibitem{schreyer2019}
M.~Schreyer, T.~Sattarov, C.~Schulze, B.~Reimer, and D.~Borth.
\newblock Detection of accounting anomalies in the latent space using
  adversarial autoencoder neural networks.
\newblock {\em 2nd KDD Workshop on Anomaly Detection in Finance, Anchorage,
  Alaska}, 2019.

\bibitem{silver2017}
D.~Silver, J.~Schrittwieser, K.~Simonyan, I.~Antonoglou, A.~Huang, A.~Guez,
  T.~Hubert, L.~Baker, M.~Lai, A.~Bolton, et~al.
\newblock Mastering the game of go without human knowledge.
\newblock {\em Nature}, 550(7676):354, 2017.

\bibitem{spathis2002}
C.~T. Spathis.
\newblock Detecting false financial statements using published data: some
  evidence from greece.
\newblock {\em Managerial Auditing Journal}, 17(4):179--191, 2002.

\bibitem{sutskever2014}
I.~Sutskever, O.~Vinyals, and Q.~V. Le.
\newblock Sequence to sequence learning with neural networks.
\newblock In {\em Advances in neural information processing systems}, pages
  3104--3112, 2014.

\bibitem{szegedy2013}
C.~Szegedy, W.~Zaremba, I.~Sutskever, J.~Bruna, D.~Erhan, I.~Goodfellow, and
  R.~Fergus.
\newblock Intriguing properties of neural networks.
\newblock {\em arXiv preprint arXiv:1312.6199}, 2013.

\bibitem{van2016}
A.~Van Den~Oord, S.~Dieleman, H.~Zen, K.~Simonyan, O.~Vinyals, A.~Graves,
  N.~Kalchbrenner, A.~W. Senior, and K.~Kavukcuoglu.
\newblock Wavenet: A generative model for raw audio.
\newblock {\em SSW}, 125, 2016.

\bibitem{wang2018}
T.-C. Wang, M.-Y. Liu, J.-Y. Zhu, A.~Tao, J.~Kautz, and B.~Catanzaro.
\newblock High-resolution image synthesis and semantic manipulation with
  conditional gans.
\newblock In {\em Proceedings of the IEEE Conference on Computer Vision and
  Pattern Recognition}, pages 8798--8807, 2018.

\bibitem{xu2015}
B.~Xu, N.~Wang, T.~Chen, and M.~Li.
\newblock {Empirical Evaluation of Rectified Activations in Convolution
  Network}.
\newblock {\em ICML Deep Learning Workshop}, pages 1--5, 2015.

\bibitem{xu2017}
X.~Xu, X.~Chen, C.~Liu, A.~Rohrbach, T.~Darell, and D.~Song.
\newblock Can you fool ai with adversarial examples on a visual turing test.
\newblock {\em arXiv preprint arXiv:1709.08693}, 2017.

\bibitem{zhou2018}
C.~Zhou, M.~Horgan, V.~Kumar, C.~Vasco, and D.~Darcy.
\newblock Voice conversion with conditional samplernn.
\newblock {\em arXiv preprint arXiv:1808.08311}, 2018.

\end{thebibliography}

\appendix
\onecolumn
\section*{Appendix A - Architectural Details}

Our architectural setup follows the AAE architecture \cite{makhzani2015}, as shown in Fig. \ref{fig:architecture}, comprised of three distinct neural networks that we trained in parallel. In the encoder network $q_{\theta}$ we use Leaky Rectified Linear Unit (LReLU) activation functions \cite{xu2015} except in the last "bottleneck" layer where we use a Hyperbolic Tangent (Tanh) activation. In both, the decoder network $p_{\theta}$ and the discriminator $d_{\phi}$ network, we use LReLUs in all layers except for the output layers. In the output layer, we use a Sigmoid (Sigm) activation. Table \ref{tab:architectures} depicts the architectural details of the networks in terms of the applied activation functions and trained number of neurons per network layer. The AAE architecture is implemented using the PyTorch machine learning library as introduced in \cite{paszke2017}.

\begin{table}[h]
    \caption{Activation function and number of neurons per layer of the distinct networks that constitute the AAE architecture: encoder $q_{\theta}$, decoder $p_{\theta}$ and discriminator $d_{\phi}$ neural network (the subscript denotes the number of neurons per layer).}
  \footnotesize
  \begin{center}
      \begin{tabular}{l|l|l}
        \multicolumn{3}{l}{\textbf{Architecture Data-A}}\\
        \multicolumn{1}{c}{} \hspace{1mm} &\multicolumn{1}{c}{}&\multicolumn{1}{c}{}\vspace{0.5mm}\\
        $x \in \mathbb{R}^{401}$ \hspace{1mm} & \hspace{1mm} $q_{\theta}(z|x)$ \hspace{1mm}& $LReLU_{256} \rightarrow LReLU_{128} \rightarrow ... \rightarrow LReLU_{16} \rightarrow LReLU_{8} \rightarrow Tanh_{2}$ \\
        $z \in \mathbb{R}^{2}$ & \hspace{1mm} $p_{\theta}(\hat{x}|z)$ & $LReLU_{8} \rightarrow LReLU_{16} \rightarrow ... \rightarrow LReLU_{128} \rightarrow LReLU_{256} \rightarrow Sigm_{401}$\\
        $z \in \mathbb{R}^{2}$& \hspace{1mm} $d_{\phi}(z)$ \hspace{1mm} & $LReLU_{128} \rightarrow LReLU_{64} \rightarrow LReLU_{32} \rightarrow LReLU_{16} \rightarrow Sigm_{1}$ \\
        \multicolumn{1}{c}{} \hspace{1mm} &\multicolumn{1}{c}{}&\multicolumn{1}{c}{}\vspace{1mm}\\
        \multicolumn{3}{l}{\textbf{Architecture Data-B}}\\
        \multicolumn{1}{c}{} \hspace{1mm} &\multicolumn{1}{c}{}&\multicolumn{1}{c}{}\vspace{0.5mm}\\
        $x \in \mathbb{R}^{618}$ \hspace{1mm} & \hspace{1mm} $q_{\theta}(z|x)$ & $LReLU_{256} \rightarrow LReLU_{64} \rightarrow LReLU_{16} \rightarrow Tanh_{2}$ \\
        $z \in \mathbb{R}^{2}$ & \hspace{1mm} $p_{\theta}(\hat{x}|z)$ & $LReLU_{16} \rightarrow LReLU_{64} \rightarrow LReLU_{256} \rightarrow Sigm_{618}$ \\
        $z \in \mathbb{R}^{2}$ & \hspace{1mm} $d_{\phi}(z)$ \hspace{1mm} & $LReLU_{256} \rightarrow LReLU_{64} \rightarrow LReLU_{16} \rightarrow Sigm_{1}$ \hspace{1mm} \\
       \end{tabular}
    \label{tab:architectures}
  \end{center}
\end{table}

\section*{Appendix B - Experimental Details}

\begin{figure*}[h!]
    \begin{center}
        \includegraphics[width=0.24\textwidth]{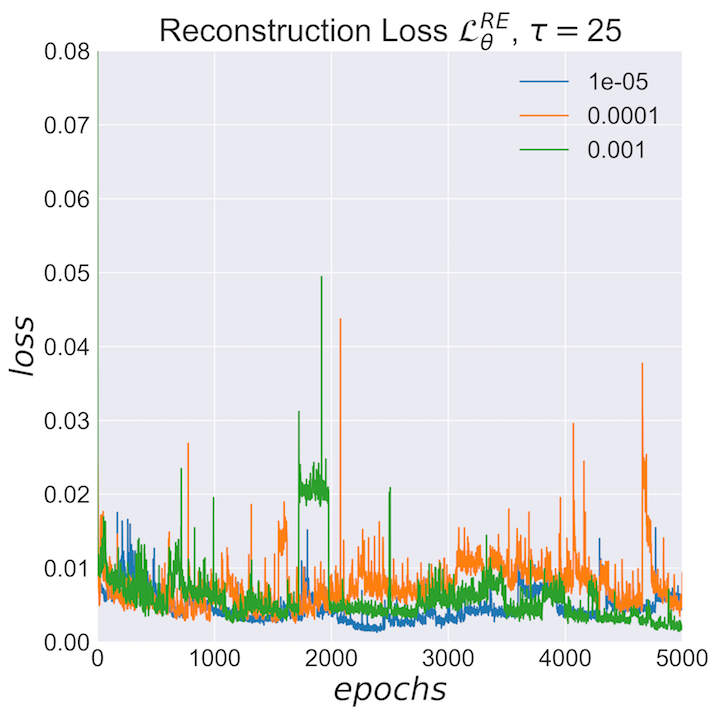}
        \includegraphics[width=0.24\textwidth]{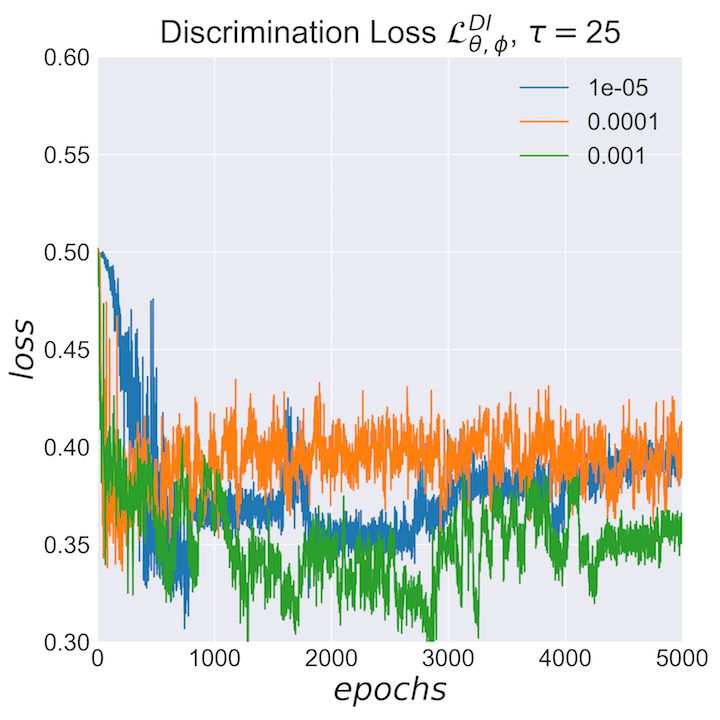}
        \includegraphics[width=0.24\textwidth]{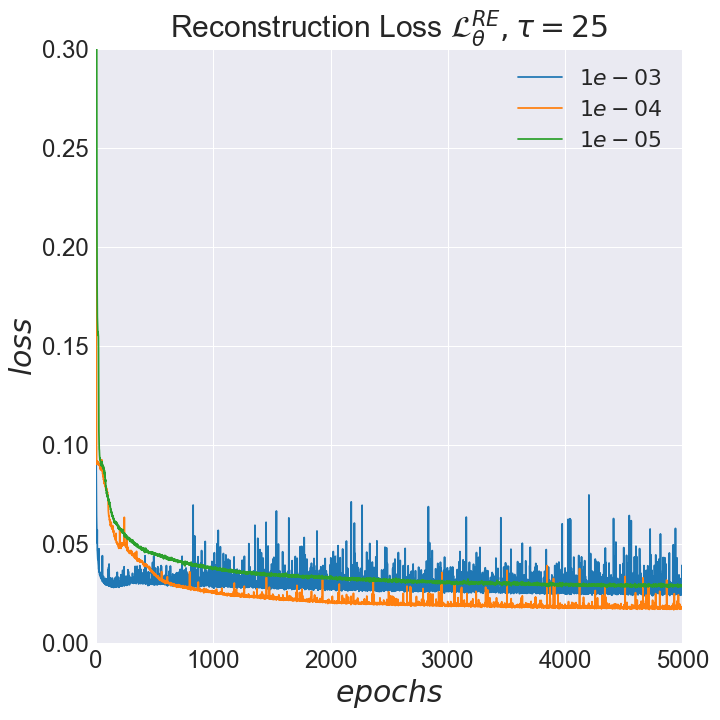}
        \includegraphics[width=0.24\textwidth]{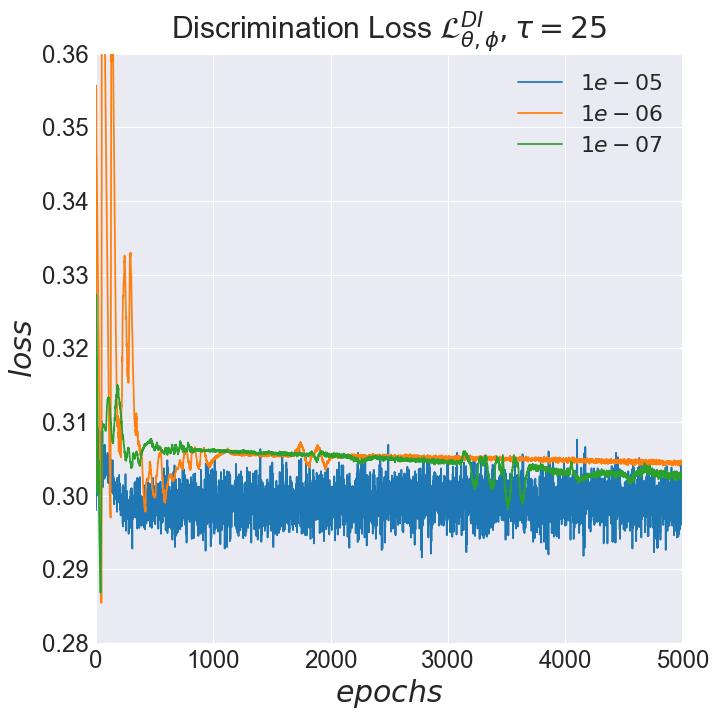}
    \end{center}
    \caption{Exemplary AAE reconstruction $\mathcal{L}^{\scaleto{RE}{3pt}}_{\theta}$ and discrimination-losses $\mathcal{L}^{\scaleto{DI}{3pt}}_{\theta, \phi}$ evaluated for Data-A (left) and Data-B (right) as well as varying learning rates $\eta$ with progressing training 5,000 epochs.}
    \label{fig:training_process}
\end{figure*}

We train each AAE architecture with mini-batch wise stochastic gradient descent for a max. of 10,000 training epochs and apply early stopping once the reconstruction loss converges. In accordance with \cite{xu2015} we set the scaling factor of the LReLUs to $\alpha = 0.4$ and initialize the AAE parameters as described in \cite{glorot2010}. A mini-batch size of 128 journal entries is used in both the reconstruction and the regularization phase. We use Adam optimization as proposed in \cite{kingma2014} and set $\beta_{1}=0.9$, $\beta_{2}=0.999$, and $\epsilon=10^{-09}$ when optimizing the network parameters. Training stability is a main challenge in adversarial training \cite{arjovsky2017} and we face a variety of collapsing and non-convergence scenarios. To determine a stable training setup we sweep the learning rates $\eta$ of the encoder and decoder networks through the interval $\eta \in [10^{-05}, 10^{-02}]$, and the learning rates of the discriminator network through the interval $\eta \in [10^{-07}, 10^{-03}]$. Ultimately, we use the following constant learning rates to learn a stable model of each dataset, Data-A: $\eta = 10^{-4}$ for the encoder and decoder, $\eta = 10^{-5}$ for the discriminator; and, Data-B: $\eta = 10^{-3}$ for the encoder and decoder, $\eta = 10^{-5}$ for the discriminator. Figure \ref{fig:training_process} illustrates the reconstruction $\mathcal{L}^{\scaleto{RE}{3pt}}_{\theta}$ and discrimination-loss $\mathcal{L}^{\scaleto{DI}{3pt}}_{\theta, \phi}$ evaluated for both datasets and varying learning rates $\eta$ with progressing training epochs.

\section*{Appendix C - High-Order Latent Factor Disentanglement}

\begin{figure*}[ht!]
    \begin{center}
    
        \includegraphics[width=0.24\textwidth]{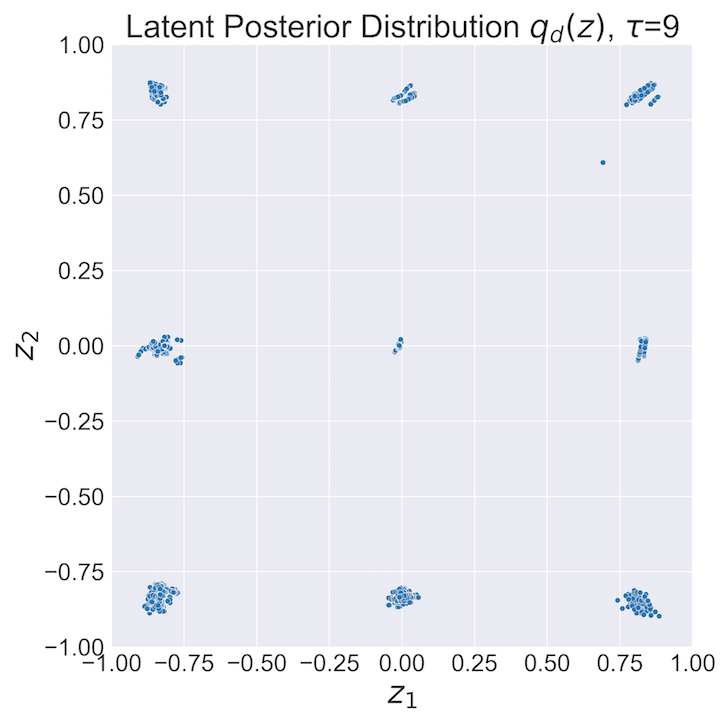}
        \includegraphics[width=0.24\textwidth]{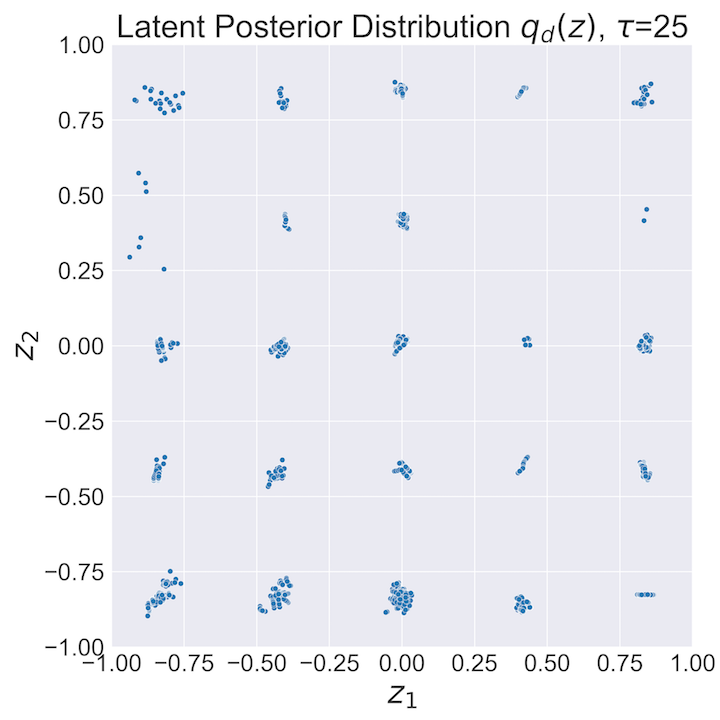}
        \includegraphics[width=0.24\textwidth]{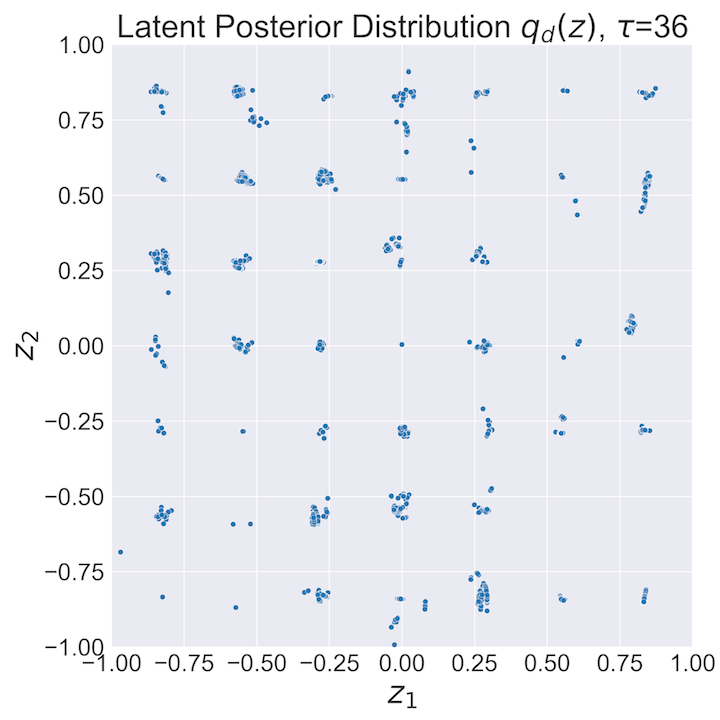}
        \includegraphics[width=0.24\textwidth]{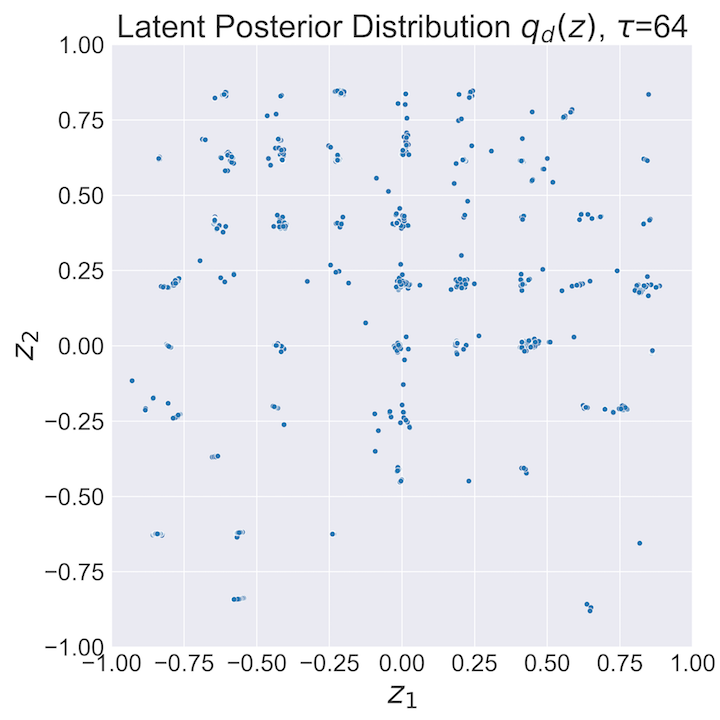}
        
        \includegraphics[width=0.24\textwidth]{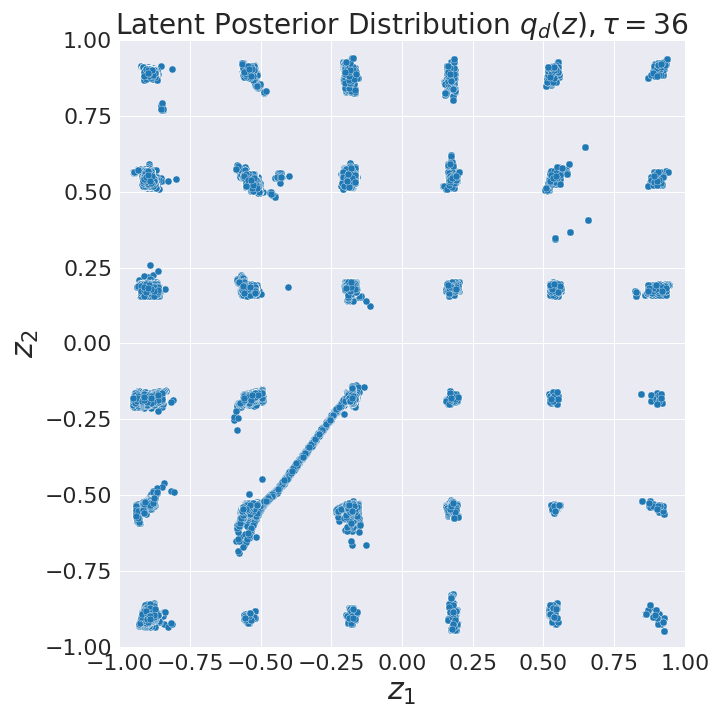}
        \includegraphics[width=0.24\textwidth]{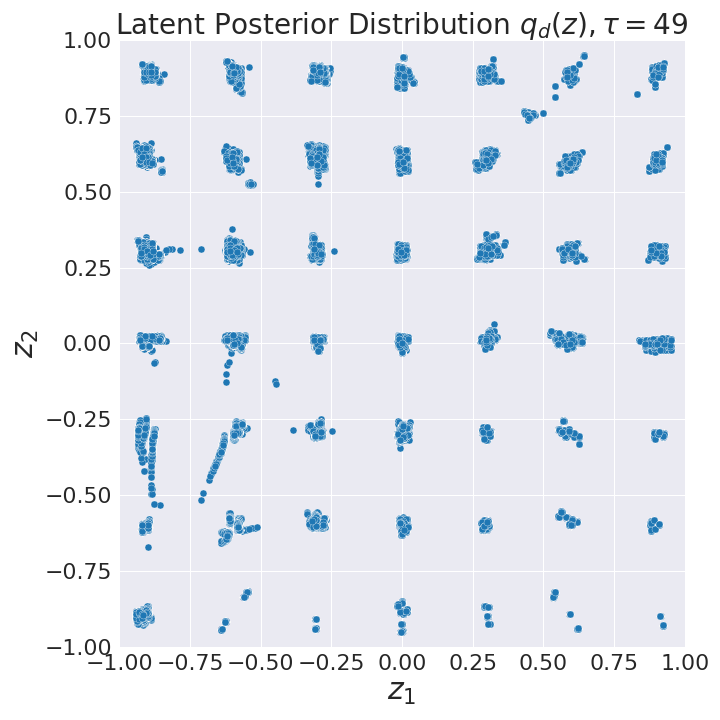}
        \includegraphics[width=0.24\textwidth]{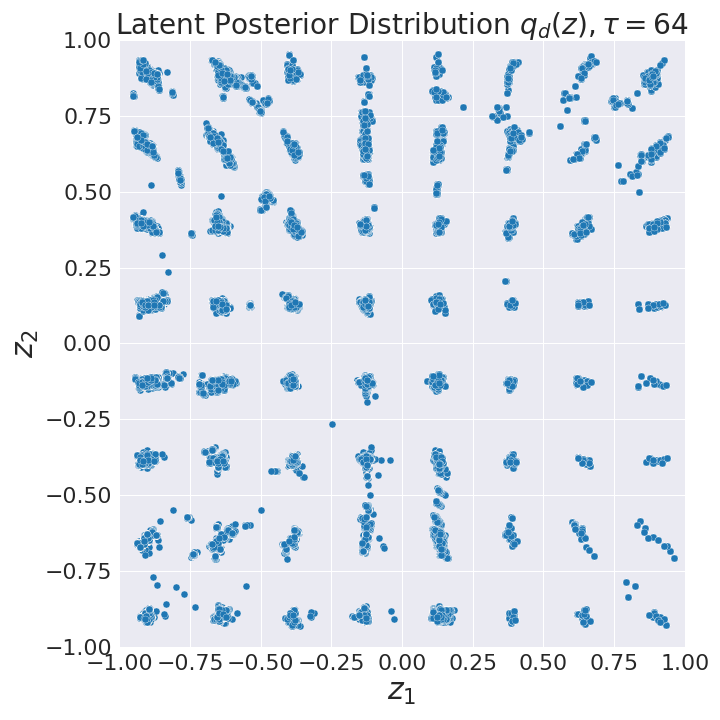}
        \includegraphics[width=0.24\textwidth]{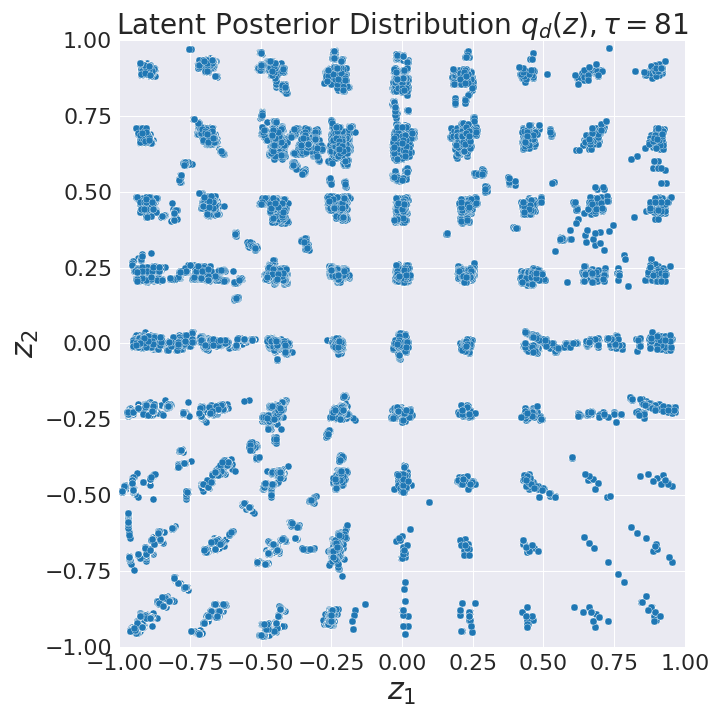}
        
    \end{center}
    \caption{Aggregated posterior distributions $q_{d}(z)$ learned when training the AAE architecture for up to 10,000 training epochs. Learned posterior $q_{d}(z)$ of Data-A when sampling from a prior $p_{n}(z)$ consisting of an equidistant 2D-grid of $\tau \in \{9, 25, 36, 64\}$ isotropic Gaussians (top row). Learned posterior $q_{d}(z)$ of Data-B when sampling from a prior $p_{n}(z)$ consisting of an equidistant 2D-grid of $\tau \in \{36, 49, 64, 81\}$ isotropic Gaussians (bottom row).}
    \label{fig:posterior_distribution_a}
\end{figure*}

To determine a semantic disentanglement of the journal entries high-order generative latent factors, we sample from a prior distribution $p_{n}(z)$ that constitutes an equidistant grid of $\tau$ multivariate isotropic Gaussians $\mathcal{N}(\mu,\mathcal{I})$, where $\mu \in \mathbb{R}^{2}$. In this initial work, we evaluate when sampling of $\tau \in \{9, 25, 36, 64\}$ Gaussians in Data-A and $\tau \in \{9, 25, 36, 49, 64, 81\}$ Gaussians in Data-B. Figure \ref{fig:posterior_distribution_a} shows the aggregated posterior distributions $q_{d}(z)$ learned when training the AAE architecture up to 10,000 training epochs on both datasets. The examination of the entries represented by each Gaussian exhibit a high semantic similarity and correspond to general generative accounting processes evident in the data, e.g., (a) automated payment runs, (b) vendor invoices, and (c) material movements. We noticed that when increasing the number $\tau$ of distinct Gaussians, each Gaussian starts to represent an even more granular accounting (sub-)process.

\section*{Appendix D - Low-Order Latent Factor Disentanglement}

To determine a semantic disentanglement of the journal entries low-order generative latent factors we obtain human interpretable sampling maps that show the results of (a) the attribute combination analysis and (b) the attribute robustness analysis in both datasets. Combining the analyses provides a 'blueprint' of controllable latent factors to generate robust adversarial journal entries. Figures \ref{fig:posterior_distribution_b} and \ref{fig:posterior_distribution_c} show exemplary robust and combination sample maps obtained of a single Gaussian $z_{k=14}$ (Data-A) and $z_{k=15}$ (Data-B) respectively. The distinct colour-levels of the combination sample maps denote the by the decoder network $p_{\theta}(\hat{x}|z)$ generated attribute value when sampling $z^{i}$ from a particular region in $Z$. The white contours denote the distinct robustness levels when sampling $z^{i}$ from a particular region in $Z$. The obtained robustness determines the likelihood of the generated entry of originating from the learned posterior $p_{d}(z)$ and therefore complying with the generative factors of the journal entries.

\begin{figure*}[ht!]
    \begin{center}
        \includegraphics[width=0.22\textwidth]{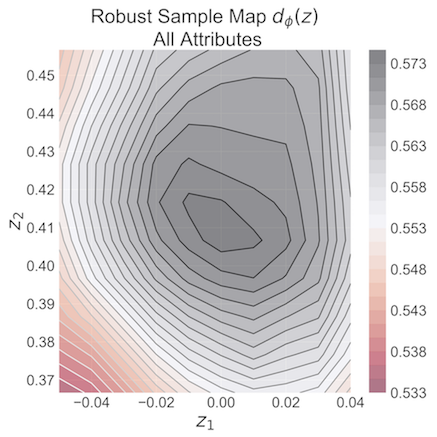}
        \includegraphics[width=0.22\textwidth]{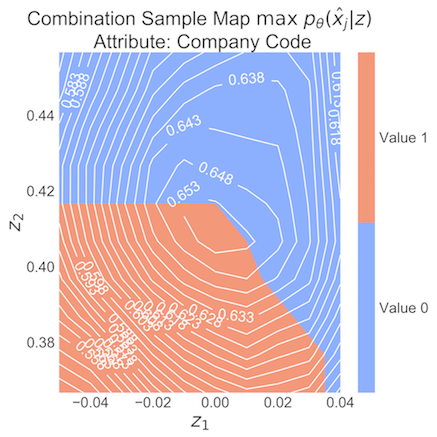}
        \includegraphics[width=0.22\textwidth]{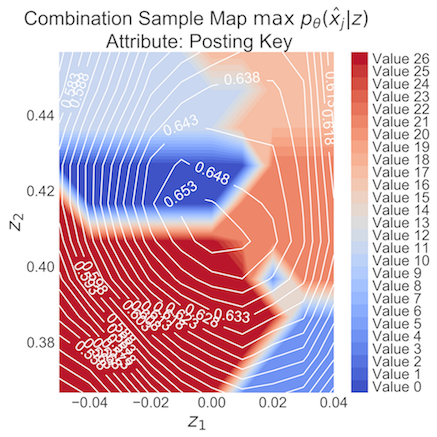}
        \includegraphics[width=0.22\textwidth]{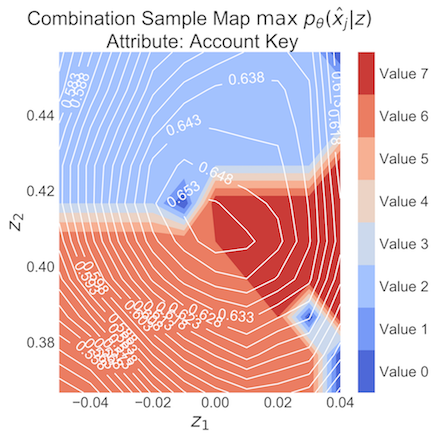}
    \end{center}
\end{figure*}

\begin{figure*}[ht!]
    \begin{center}
        \includegraphics[width=0.22\textwidth]{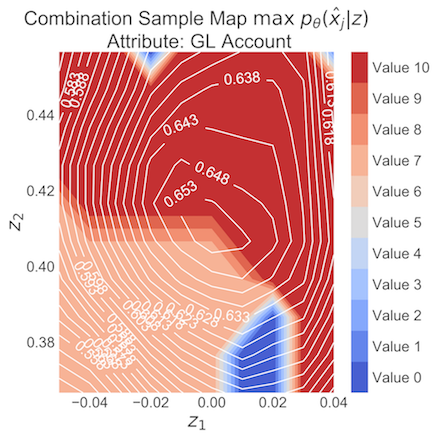}
        \includegraphics[width=0.22\textwidth]{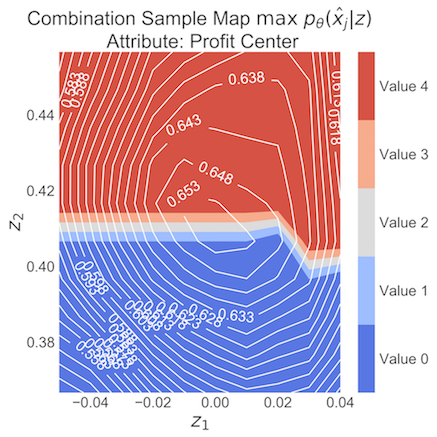}
        \includegraphics[width=0.22\textwidth]{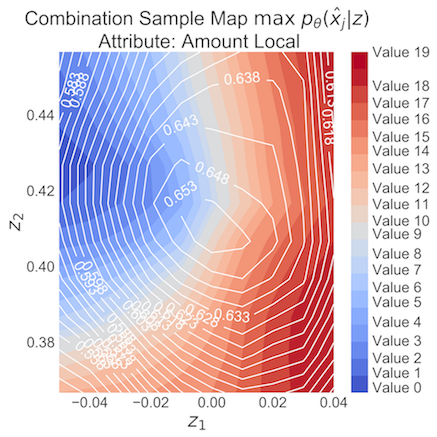}
        \includegraphics[width=0.22\textwidth]{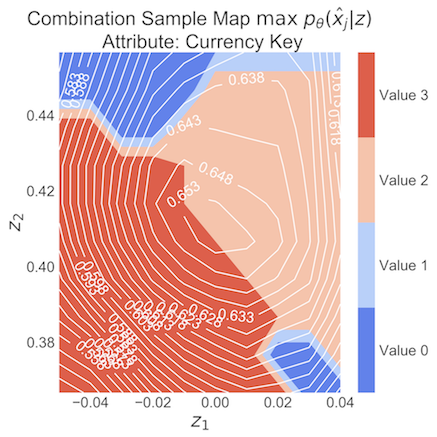}
    \end{center}
    \caption{Exemplary robust sample map obtained for Gaussian $z_{k=14}$ in Data-A (top-left) and corresponding combination sample maps of the distinct journal entry attributes $x_{j}$ (others). The corresponding AAE model is trained for 10,000 training epochs imposing a prior comprised of $\tau=25$ equidistant isotropic Gaussians.}
    \label{fig:posterior_distribution_b}
\end{figure*}

\begin{figure*}[ht!]
    \begin{center}
        \includegraphics[width=0.22\textwidth]{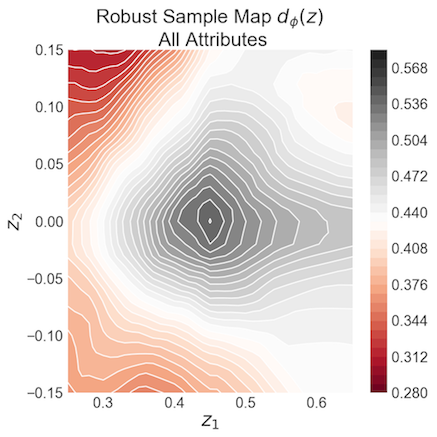}
        \includegraphics[width=0.22\textwidth]{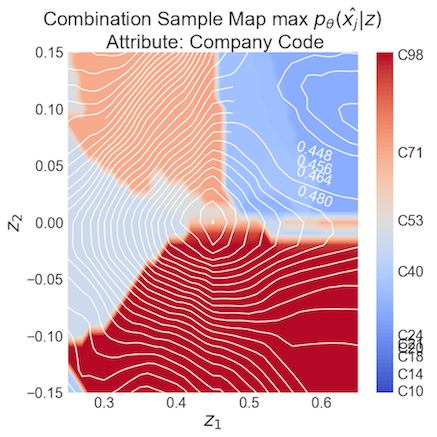}
        \includegraphics[width=0.22\textwidth]{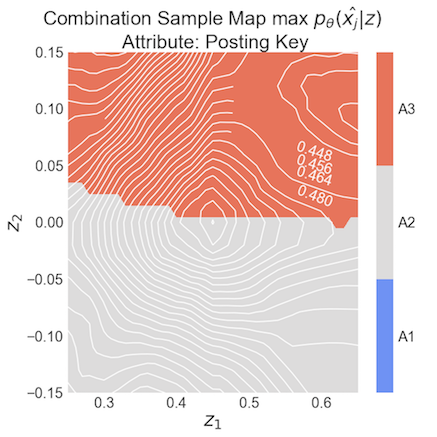}
        \includegraphics[width=0.22\textwidth]{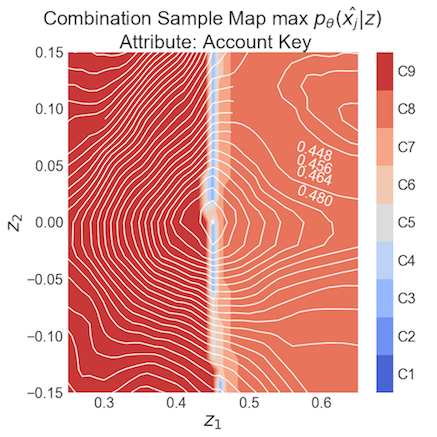}
    \end{center}
\end{figure*}

\begin{figure*}[ht!]
    \begin{center}
        \includegraphics[width=0.22\textwidth]{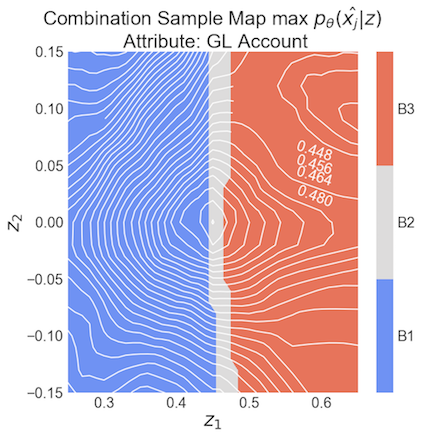}
        \includegraphics[width=0.22\textwidth]{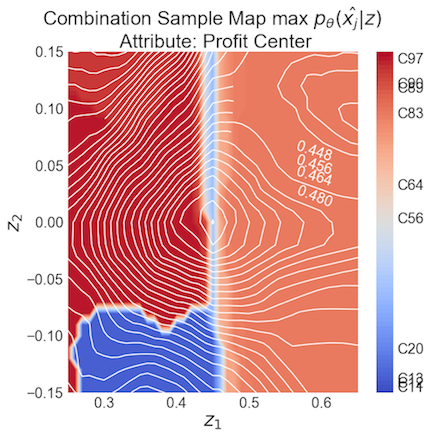}
        \includegraphics[width=0.22\textwidth]{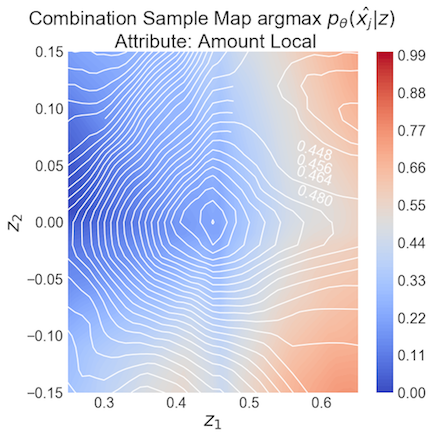}
        \includegraphics[width=0.22\textwidth]{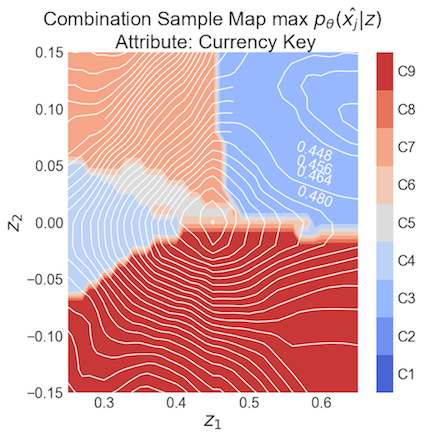}
    \end{center}
    \caption{Exemplary robust sample map obtained for Gaussian $z_{k=15}$ in Data-B (top-left) and corresponding combination sample maps of the distinct journal entry attributes $x_{j}$ (others). The corresponding AAE model is trained for 10,000 training epochs imposing a prior comprised of $\tau=25$ equidistant isotropic Gaussians.}
    \label{fig:posterior_distribution_c}
\end{figure*}

\newpage

\section*{Appendix E - Robust Sampling of Adversarial Journal Entries}

Figure \ref{fig:posterior_traversal_a} illustrates an exemplary traversal in the adversarial sampling region $q_{s}(z_{k=15})$ learned from Data-B. The sampling region is determined by $d_{\phi}(z^{i}) \geq 0.49$ and is conducted equidistant in $z_{1} \in [-0.2; 0.6]$ while conditioning on $z_{2}=0$ using an Euclidean distance of $\delta=0.02$ between two neighboring samples $z^{i}$ and $z^{l}$ with $i \neq l$. Given the learned AAE's decoder $p_{\theta}(\hat{x}|z)$, each sample $z^{i} \in Z$ generates an adversarial journal entry $X_{Adv}$ exhibiting the attribute values illustrated in the distinct combination sample maps of Fig. \ref{fig:posterior_distribution_c}. It can be observed that traversing the $z_{1}$-dimension controls (among other attributes) the local posting amount of the by the decoder generated entries shown in Tab. \ref{tab:sampled_entries}. Traversing towards the mode of the adversarial sampling region results in an increased $d_{\phi}(z)$ robustness of the generated journal entries against the CAATS usually applied by auditors. Hence, the likelihood of being generated by the same latent factor distributions as the regular journal entry population increases. Figure \ref{fig:posterior_traversal_b} illustrates a random sampling outside of the same adversarial sampling region. It can be observed that traversing away from the adversarial sampling region results in journal entries that exhibit an 'unusual' high or low posting amount. Hence, the likelihood of being generated by the same latent factor distributions as the regular journal entry population decreases.

\begin{figure*}[!ht]
    \begin{center}
        \includegraphics[width=0.22\textwidth]{02b_latent_space_sampling_disc_loss_mode_14.png}
        \includegraphics[width=0.22\textwidth]{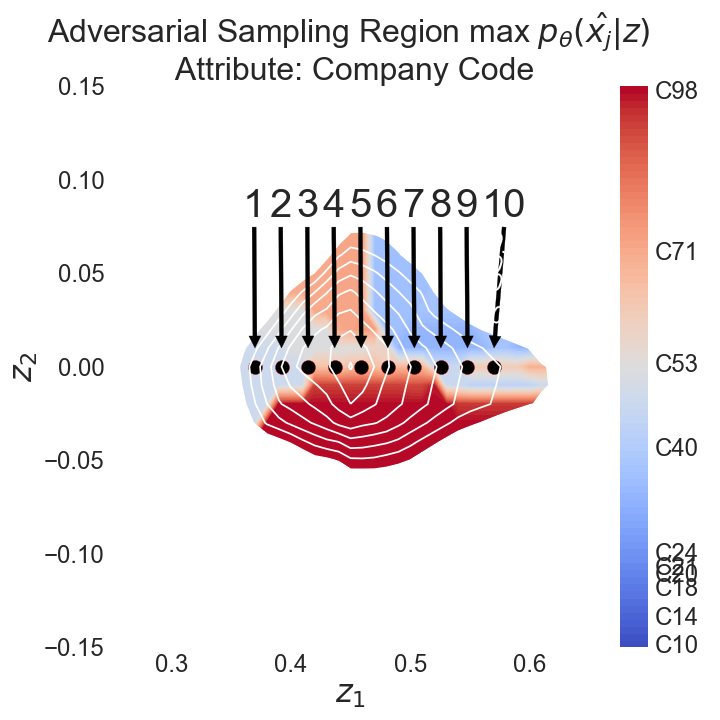}
        \includegraphics[width=0.22\textwidth]{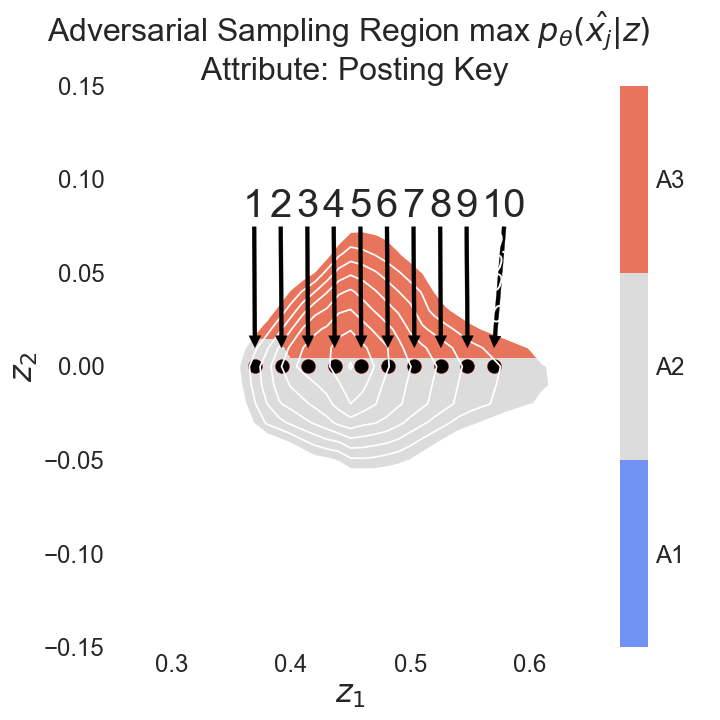}
        \includegraphics[width=0.22\textwidth]{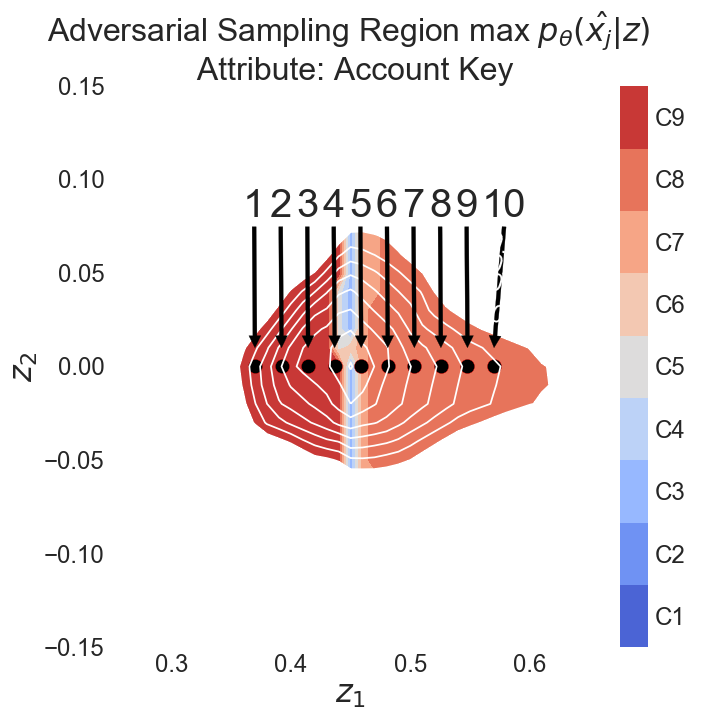}
    \end{center}
\end{figure*}
    
\begin{figure*}[h!]
    \begin{center}
        \includegraphics[width=0.22\textwidth]{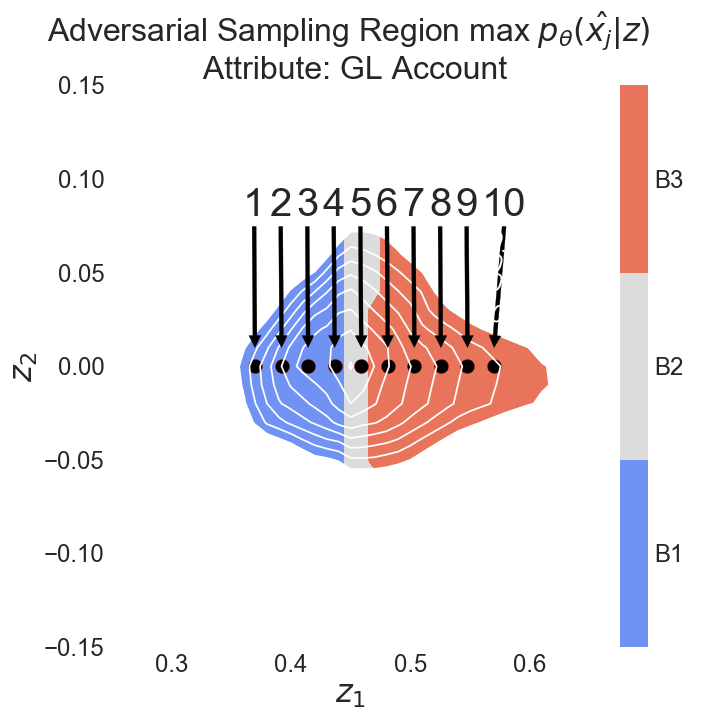}
        \includegraphics[width=0.22\textwidth]{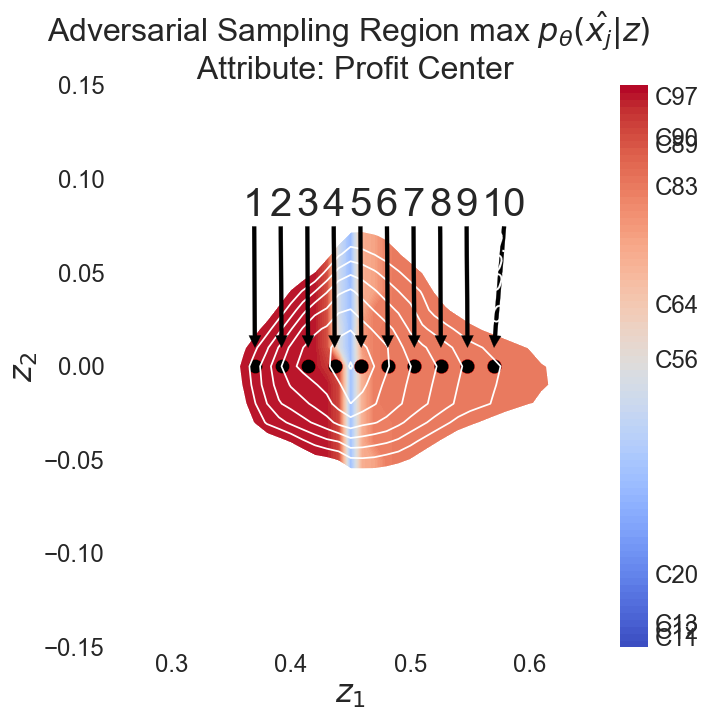}
        \includegraphics[width=0.22\textwidth]{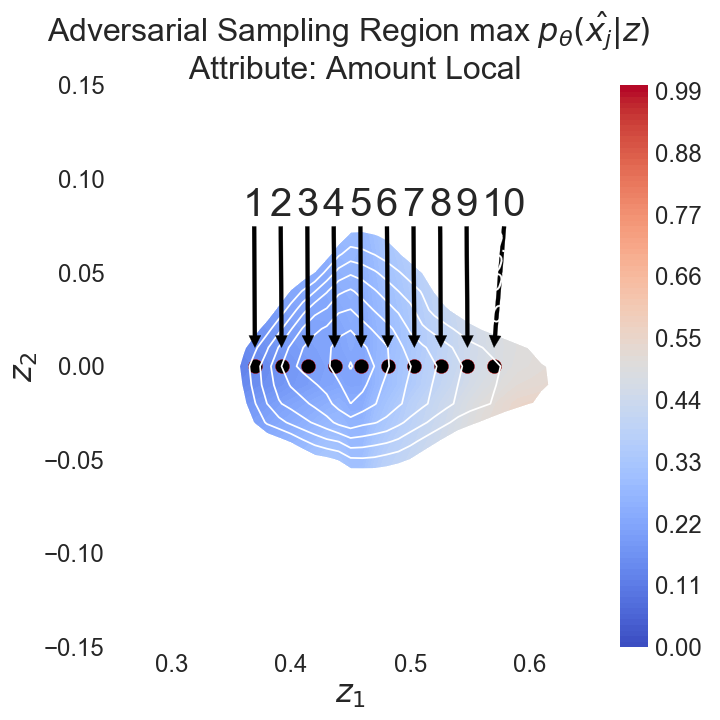}
        \includegraphics[width=0.22\textwidth]{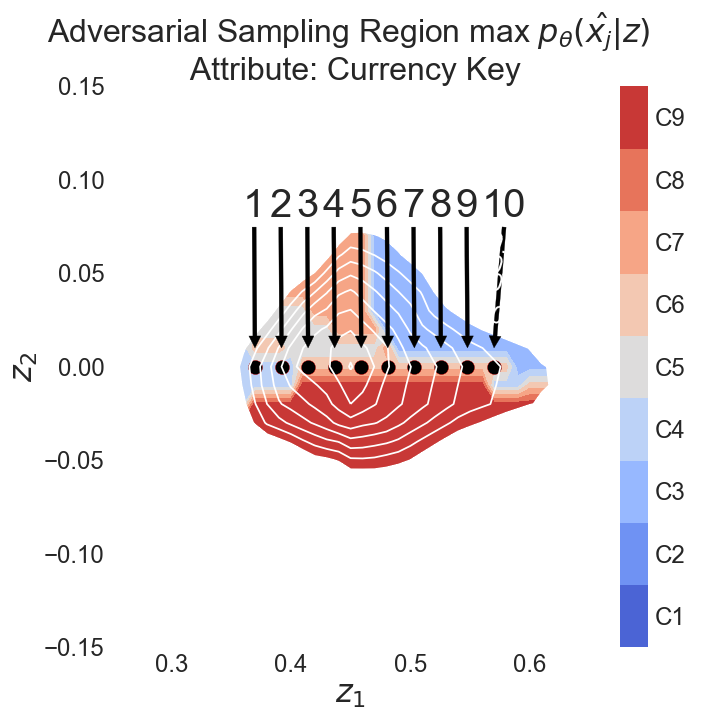}
    \end{center}
    \caption{Exemplary equidistant traversal in $z_{1} \in [-0.2; 0.6]$ with $\delta=0.02$ while keeping $z_{2}=0$ fixed in the adversarial sampling region $q_{s}(z_{k=15})$ of Data-B. Sample ids $i$ are denoted by the black arrows of the adversarial sampling maps. The individual journal entries generated per sample $z^{i}$ in $q_{s}(z_{k})$ are shown in Tab. \ref{tab:sampled_entries}.}
    \label{fig:posterior_traversal_a}
\end{figure*}

\begin{figure*}[!ht]
    \begin{center}
        \includegraphics[width=0.22\textwidth]{02b_latent_space_sampling_disc_loss_mode_14.png}
        \includegraphics[width=0.22\textwidth]{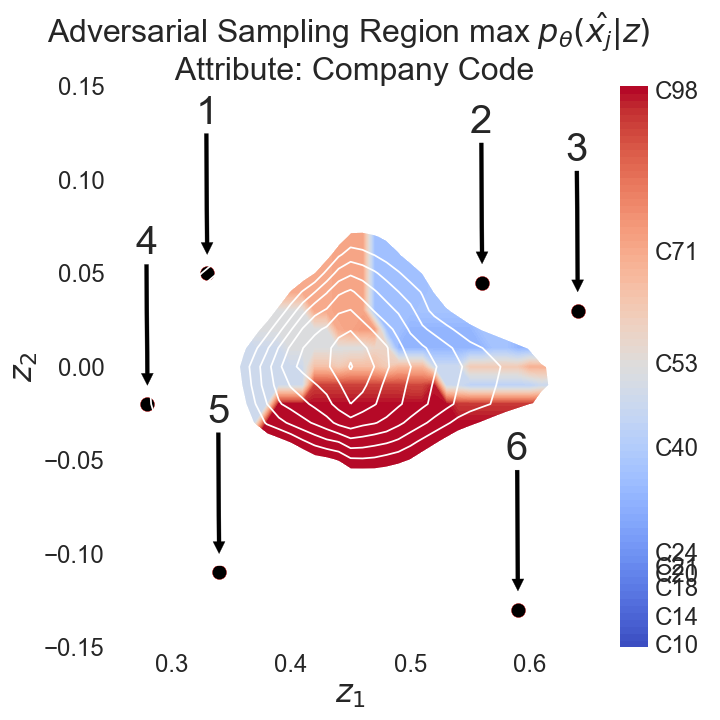}
        \includegraphics[width=0.22\textwidth]{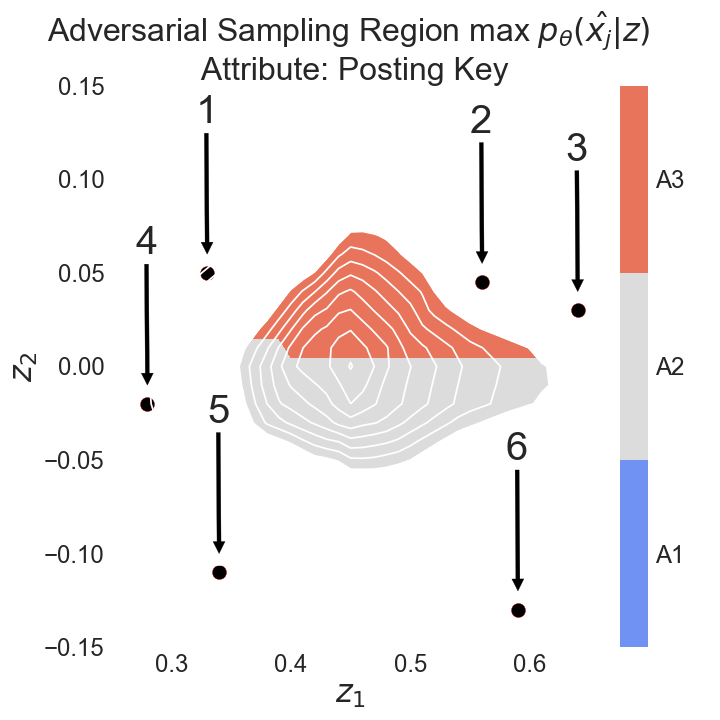}
        \includegraphics[width=0.22\textwidth]{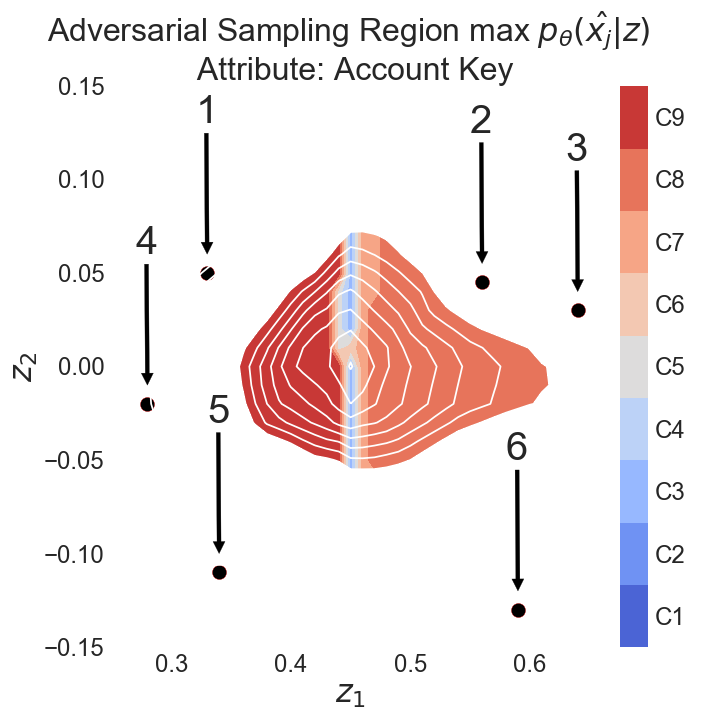}
    \end{center}
\end{figure*}
    
\begin{figure*}[h!]
    \begin{center}
        \includegraphics[width=0.22\textwidth]{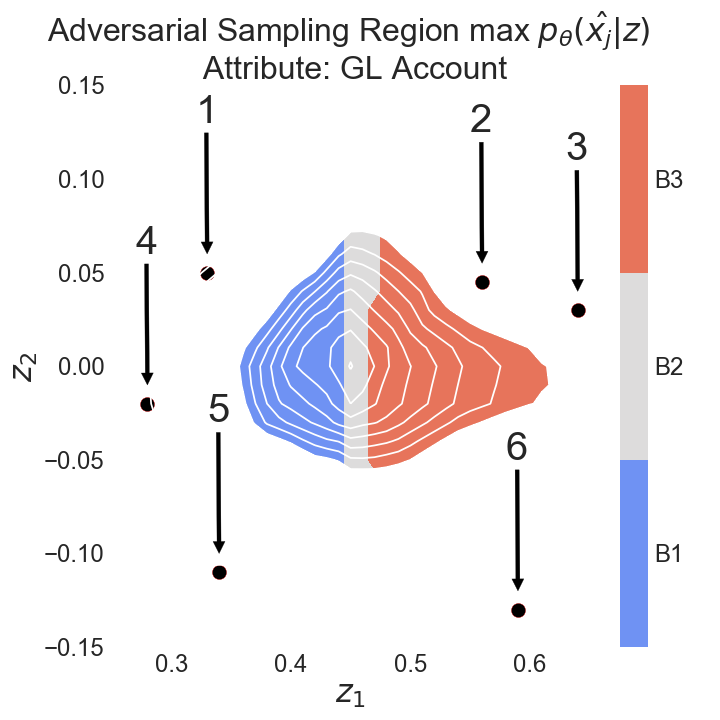}
        \includegraphics[width=0.22\textwidth]{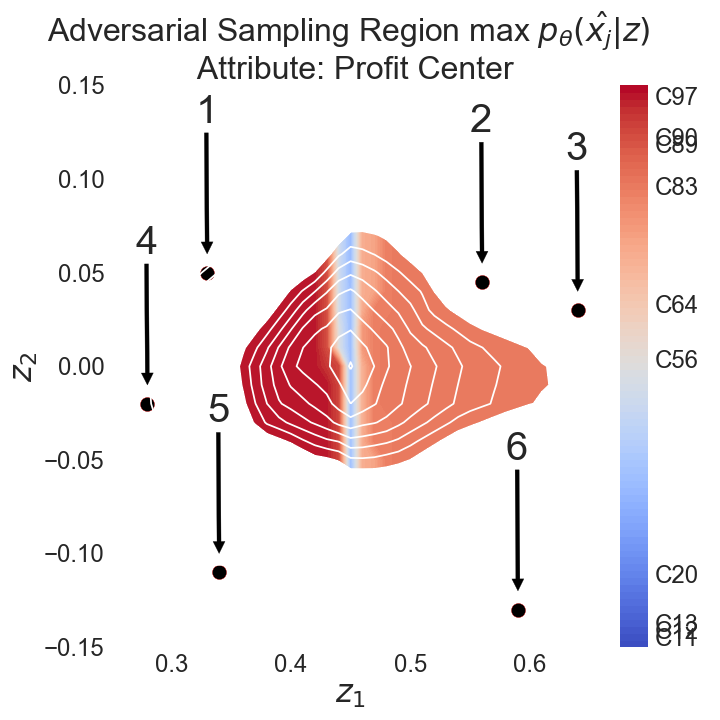}
        \includegraphics[width=0.22\textwidth]{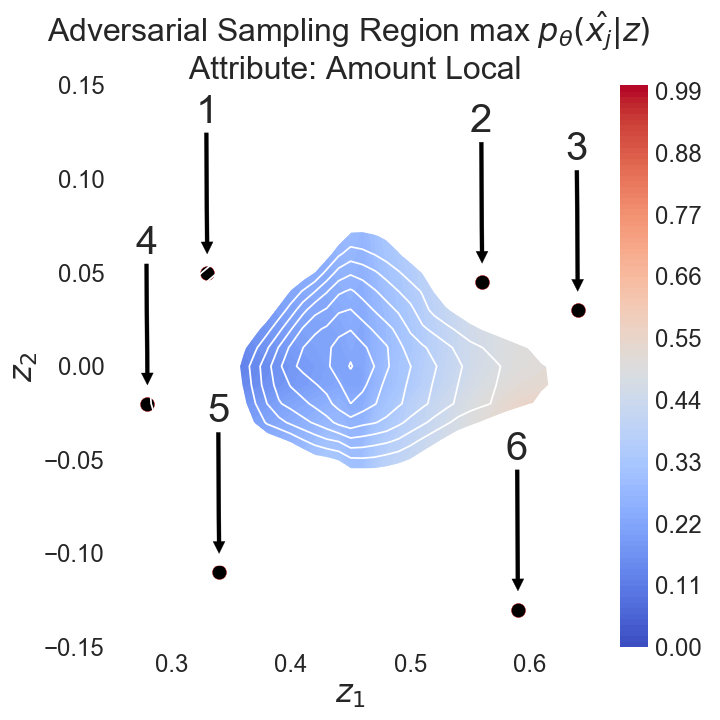}
        \includegraphics[width=0.22\textwidth]{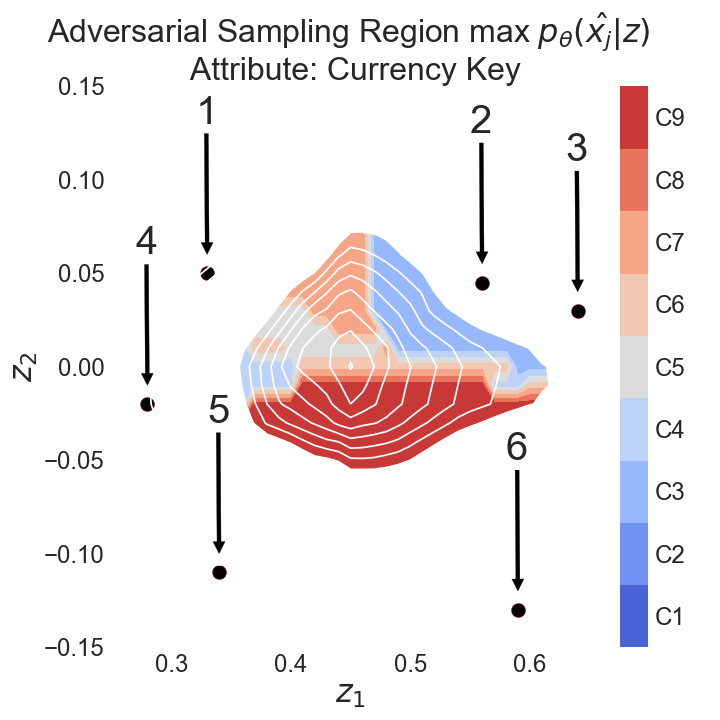}
    \end{center}
    \caption{Exemplary random sampling outside of the adversarial sampling region $q_{s}(z_{k=15})$ in $z_{1} \in [-0.2; 0.6]$ and $z_{2} \in [0.15; -0.15]$ of Data-B. Sample ids $i$ are denoted by the black arrows of the adversarial sampling maps. The individual journal entries generated per sample $z^{i}$ in $q_{s}(z_{k})$ are shown in Tab. \ref{tab:sampled_entries_fake}}
    \label{fig:posterior_traversal_b}
\end{figure*}

\begin{table}[!ht]
    \caption{Generated journal entries of each sample id $i$ when equidistant traversing the $z_{1}$-dimension in $q_{s}(z_{k=15})$ with $\delta=0.02$ and corresponding adversarial robustness $d_{\phi}(z)$. Traversing away from the sampling region mode results in a likelihood increase of not being generated by the same latent factor distributions as the regular journal entry population. As a result, the likelihood of an alert triggered by a CAATs increases.}
    \label{tab:sampled_entries}
  \scriptsize
  \begin{center}
    \begin{tabular}{c |c|c|c|c|c|r|c|c c c c}
         \multicolumn{1}{c}{\textbf{  }}
        & \multicolumn{1}{c}{\textbf{ Company }}
        & \multicolumn{1}{c}{\textbf{ Posting }} 
        & \multicolumn{1}{c}{\textbf{ Account }}
        & \multicolumn{1}{c}{\textbf{ GL }}
        & \multicolumn{1}{c}{\textbf{ Profit }}
        & \multicolumn{1}{c}{\textbf{ Amount }}
        & \multicolumn{1}{c}{\textbf{ ... }}
        & \multicolumn{1}{c}{\textbf{ Currency }} 
        & \multicolumn{1}{c}{\textbf{  }}
        & \multicolumn{1}{r}{\textbf{ }}\\
         \multicolumn{1}{c}{\textbf{  }}
        & \multicolumn{1}{c}{\textbf{ Code }}
        & \multicolumn{1}{c}{\textbf{ Key }} 
        & \multicolumn{1}{c}{\textbf{ Key }}
        & \multicolumn{1}{c}{\textbf{ Account }}
        & \multicolumn{1}{c}{\textbf{ Center }}
        & \multicolumn{1}{c}{\textbf{ Local }}
        & \multicolumn{1}{c}{\textbf{ }}
        & \multicolumn{1}{c}{\textbf{ Key }} 
        & \multicolumn{1}{c}{\textbf{  }}
        & \multicolumn{1}{r}{\textbf{$d_{\phi}(z^{i})$}} \\
        \cmidrule{1-9} \cmidrule{11-11} \morecmidrules\cmidrule{1-9} \cmidrule{11-11}
         1 & C48 & A2 & C9 & B1 & C97 & \textbf{125.68} & ... & C4 &  & 0.50\\
         2 & C48 & A2 & C9 & B1 & C97 & \textbf{165.75} & ... & C4 &  & 0.52\\
         3 & C48 & A2 & C9 & B1 & C97 & \textbf{214.86} & ... & C6 &  & 0.53\\
         4 & C64 & A2 & C9 & B1 & C91 & \textbf{271.46} & ... & C6 &  & 0.53\\
         5 & C65 & A2 & C7 & B2 & C71 & \textbf{262.67} & ... & C6 &  & 0.54\\
         6 & C61 & A2 & C8 & B3 & C83 & \textbf{625.61} & ... & C6 &  & 0.52\\
         7 & C66 & A2 & C8 & B3 & C83 & \textbf{1078.08} & ... & C6 &  & 0.51\\
         8 & C47 & A2 & C8 & B3 & C83 & \textbf{1796.61} & ... & C6 &  & 0.51\\
         9 & C47 & A2 & C8 & B3 & C83 & \textbf{2899.71} & ... & C6 &  & 0.50\\
         10 & C61 & A2 & C8 & B3 & C83 & \textbf{4095.49} & ... & C6 &  & 0.49\\
    \end{tabular}
\end{center}
\end{table}

\begin{table}[!ht]
    \caption{Generated journal entries of each sample id $i$ when sampling randomly outside of the adversarial sampling region $q_{s}(z_{k=15})$ with $d_{\phi}(z) \leq 0.49$ and corresponding adversarial robustness $d_{\phi}(z)$. Traversing away from the sampling region mode results in a likelihood increase of not being generated by the same latent factor distributions as the regular journal entry population. As a result, the likelihood of an alert triggered by a CAATs increases.}
    \label{tab:sampled_entries_fake}
  \scriptsize
  \begin{center}
    \begin{tabular}{c |c|c|c|c|c|r|c|c c c c}
         \multicolumn{1}{c}{\textbf{  }}
        & \multicolumn{1}{c}{\textbf{ Company }}
        & \multicolumn{1}{c}{\textbf{ Posting }} 
        & \multicolumn{1}{c}{\textbf{ Account }}
        & \multicolumn{1}{c}{\textbf{ GL }}
        & \multicolumn{1}{c}{\textbf{ Profit }}
        & \multicolumn{1}{c}{\textbf{ Amount }}
        & \multicolumn{1}{c}{\textbf{ ... }}
        & \multicolumn{1}{c}{\textbf{ Currency }} 
        & \multicolumn{1}{c}{\textbf{  }}
        & \multicolumn{1}{r}{\textbf{ }}\\
         \multicolumn{1}{c}{\textbf{  }}
        & \multicolumn{1}{c}{\textbf{ Code }}
        & \multicolumn{1}{c}{\textbf{ Key }} 
        & \multicolumn{1}{c}{\textbf{ Key }}
        & \multicolumn{1}{c}{\textbf{ Account }}
        & \multicolumn{1}{c}{\textbf{ Center }}
        & \multicolumn{1}{c}{\textbf{ Local }}
        & \multicolumn{1}{c}{\textbf{ }}
        & \multicolumn{1}{c}{\textbf{ Key }} 
        & \multicolumn{1}{c}{\textbf{  }}
        & \multicolumn{1}{r}{\textbf{$d_{\phi}(z^{i})$}}\\
        \cmidrule{1-9} \cmidrule{11-11} \morecmidrules\cmidrule{1-9} \cmidrule{11-11}
         1 & C71 & A3 & C9 & B1 & C97 & \textbf{134.28} & ... & C5 &  & 0.42\\
         2 & C37 & A3 & C8 & B3 & C83 & \textbf{4236.12} & ... & C3 &  & 0.46\\
         3 & C30 & A3 & C8 & B3 & C83 & \textbf{10043.45} & ... & C3 &  & 0.46\\
         4 & C48 & A2 & C9 & B1 & C97 & \textbf{93.65} & ... & C4 &  & 0.42\\
         5 & C98 & A2 & C8 & B1 & C13 & \textbf{1604.02} & ... & C9 &  & 0.38\\
         6 & C98 & A2 & C8 & B3 & C83 & \textbf{34278.40} & ... & C9 &  & 0.43\\
    \end{tabular}
\end{center}
\end{table}

\break

\textcolor{white}{margin}

\end{document}